\documentclass[conference]{IEEEtran}
\usepackage{cite}
\usepackage{amsmath,amssymb,amsfonts}
\usepackage{algorithmic}
\usepackage{graphicx}
\usepackage{textcomp}
\usepackage{xcolor}
\def\BibTeX{{\rm B\kern-.05em{\sc i\kern-.025em b}\kern-.08em
    T\kern-.1667em\lower.7ex\hbox{E}\kern-.125emX}}

\usepackage{tikz}
\usetikzlibrary{arrows,calc,shapes,decorations.pathreplacing}
\usepackage{pgfplots}
\usepackage{paralist}

\usepackage{adjustbox}
\usepackage{subfig}
\usepackage{multirow}

\newcommand{\rv}[1]{\widehat{#1}}
\newcommand{\fatR}{\mathbb{R}}
\newcommand{\bsection}[1]{\section{\textbf{#1}}}
\newcommand{\bsubsection}[1]{\subsection{\textbf{#1}}}
\newcommand{\bsubsectionS}[1]{\subsection*{\textbf{#1}}}

\newcommand{\reff}[1]{(\ref{#1})}
\newcommand{\Ereff}[1]{Equation~\reff{#1}}
\newcommand{\fR}[1]{\mathbb{R}^{#1}}
\newcommand{\ex}[1]{E\{ #1 \}}

\newcommand{\dc}{\text{$d_\text{classic}$}}
\newcommand{\de}{\text{$d_\text{ent}$}}
\newcommand{\dl}{\text{$d_\text{lower}$}}

\newcommand{\rc}{\text{$r_\text{classic}$}}
\newcommand{\re}{\text{$r_\text{ent}$}}

\begin{document}

\title{
Improving the Accuracy of Principal Component Analysis
by the
Maximum Entropy Method
}

\author{
  \IEEEauthorblockN{Guihong Wan}
  \IEEEauthorblockA{\textit{Department of Computer Science} \\
    \textit{The University of Texas at Dallas}\\
    Richardson, Texas 75083\\
    Guihong.Wan@utdallas.edu}
  \and
  \IEEEauthorblockN{Crystal Maung}
  \IEEEauthorblockA{\textit{7 Next} \\
    \textit{7-Eleven Inc.}\\
    Irving, Texas 75063\\
    Crystal.Maung@7-11.com}
  \and
  \IEEEauthorblockN{Haim Schweitzer}
  \IEEEauthorblockA{\textit{Department of Computer Science} \\
    \textit{The University of Texas at Dallas}\\
    Richardson, Texas 75083\\
    HSchweitzer@utdallas.edu}
}

%
%%%%%%%%%%%%%%%%%%%%%%%%%%%%%%%%%%%%%%%%%%%%%%%%%%

\maketitle

\begin{abstract}
Classical Principal Component Analysis (PCA) approximates data 
in terms of projections on a small number of orthogonal vectors.
There are simple procedures to
efficiently compute various functions of the data
from the PCA approximation.
The most important function is arguably 
the Euclidean distance between data items,
This can be used, for example,
to solve the approximate nearest neighbor problem.
We use random variables to model
the inherent uncertainty in such approximations,
and apply the Maximum Entropy Method
to infer the underlying probability distribution.
We propose using the expected values of distances between these
random  variables as improved estimates of the distance.
We show by analysis and experimentally that in most cases
results obtained by our method
are more accurate than what is obtained by the classical approach.
This improves the accuracy of a classical technique
that have been used with little change for over 100 years.
\end{abstract}


\bsection{Introduction} \label{secIntro}

%%%%%%%%%%%%%%%%%%%%%%%%%%%%%%%%%%%%%%%%%%%%%%%%%%

\begin{figure}
\begin{center}
  \begin{tikzpicture}[scale=1.5]
\draw [blue,thick] (0,0) rectangle (4,2);
\draw [decorate,
        decoration={brace,amplitude=10pt},
        xshift=-4pt,yshift=0pt]
        (0,0) -- (0,2) node [black,midway,xshift=-0.6cm]
        {\footnotesize $m$};
\draw [decorate,
        decoration={brace,amplitude=10pt},
        xshift=0pt,yshift=4pt]
        (0,2) -- (4,2) node [black,midway,yshift=0.6cm]
        {\footnotesize $n$};

\draw [green,thick] (0,0.5) rectangle (4,0.7);
\node at (3.9,0.6) (A) {};
\node (B) [right of=A] {};
\node (C) [above of=B] {a row};
\node (D) [below of=C] {a column};
\draw [->] (C) -- (A);
\draw [red,thick] (3,0) rectangle (3.1,2);
\draw [->] (D) .. controls (4,-0.5) .. (3.05,0);
\end{tikzpicture}
  \end{center}
\caption{
The view of data as a matrix.
There are $n$ data items, and each one is of size $m$.
A data item is a column of an $m \times n$ matrix.
}
\label{figmn}
\end{figure}
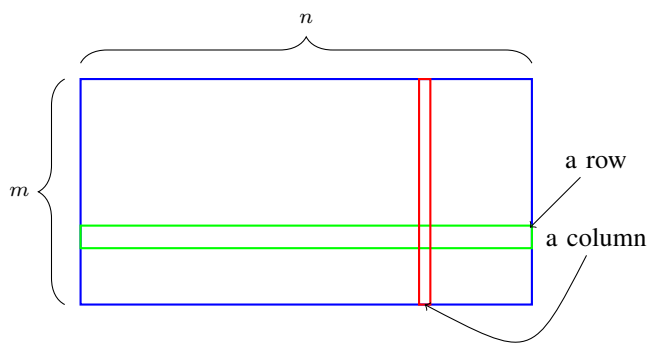

We consider the standard representation of numerical data as 
a large matrix of numeric values.
Let $n$ be the number of data items in the dataset, and let $m$ be the size
of each item.
The data can be viewed as a matrix of size $m \times n$,
as illustrated in Fig.~\ref{figmn}.
In many practical situations both $m$ and $n$ are very large.
For example, datasets containing genome data
may have $m$ in the thousands and $n$ in the millions~\cite{HapMap05}.
In such cases even simple tasks, such as searching the data
for a particular item become computationally expensive.

A standard approach to address this ``curse of dimensionality''
is dimension reduction,
reducing the dimension of each data item from $m$ to $k$, where $k < m$.
For a review of dimension reduction techniques 
see, e.g.,~\cite{Burges10:book}.
The most common approach is the Principal Component Analysis (PCA),
known for over 100 years.
For references see,
for example~\cite{Gray17,Jolliffe02,HLR:AAAI19,astarRPCA:IJAIT}.
The uncentered variant
can be described as follows.
Let $A$ be the data matrix of size $m \times n$;
define the $m \times m$ matrix $B$ by: $B = A A^T$.
Let $V$ be an $m \times k$ matrix whose columns are the
$k$ eigenvectors of $B$ corresponding to the $k$ largest eigenvalues.
The columns of $V$ are orthogonal, 
and their span gives the best possible
approximation of rank $k$ to the column space of $A$.
Let $a_i$ be the $i$th column of $A$.
The following approximations hold:
\begin{equation} \label{AVW}
  A \approx V W, \quad  a_i \approx V w_i
\end{equation}
Here $W$ is $k \times n$, representing $A$ in the reduced dimension.
In particular, the $i$th column of $A$ is the vector $a_i$, 
and it is represented by $w_i$, the $i$th column of $W$.
The matrix $W$ or any specific column $w_i$ can be computed by:
\begin{equation} \label{WVTA}
W = V^T A, \quad w_i = V^T a_i
\end{equation}
The centered variant of the PCA is the same as the uncentered PCA
with an initial centering of each column.
The centering is performed by mean subtraction.
See, e.g.,~\cite{Cadima09}.
The PCA enables fast computations of many data related techniques.
The low dimension also helps with
the visualization and the interpretation of the data.

We proceed to describe 
how to use the representation in \reff{AVW}
to approximate the Euclidean  distance between data items.
Recall that the squared Euclidean distance between two vectors $x$ and $y$
is given by:
\[
\text{distance}^2(x,y) = \|x - y\|^2
\]
It is
computed as the sum of $m$ squared coordinates.
Thus, the cost of computing this distance is $O(m)$.
Now suppose both $x$ and $y$ are from the same dataset
with known PCA,
as given by \Ereff{AVW}.
For clarity we take $x=a_i$ and $y=a_j$.
Then because of the orthogonality of $V$ we have:
\begin{multline} \label{aiaj}
\text{distance}^2(a_i,a_j) = \| a_i - a_j \|^2
\\
\approx \| V (w_i - w_j) \|^2 = \| w_i - w_j \|^2
\end{multline}
This is a classical approximation formula, known for over 100 years.
See, e.g., \cite{Jolliffe02,Burges10:book}.
It shows that the approximate value of $\| a_i - a_j \|^2$
can be computed in $O(k)$ instead of the exact computation which
takes $O(m)$.

Another common situation where the PCA leads to
significant improvements in the running time is the following.
Suppose the vector $x$ is not necessarily a column of $A$,
and one has to calculate the $n$ squared distances
$d_i^2 = \| x - a_i \|^2$ for $i=1,\ldots, n$.
(These are the distances between $x$ and all the columns of $A$.)
This situation occurs, for example,
in calculating the nearest neighbor of $x$ among the columns of $A$
(e.g., \cite{Weber98}),
or  in the computation of multi-dimensional scaling
(e.g., \cite{CoxCox}).
The direct approach requires computing $n$ distances which takes $O(mn)$.
If the PCA of $A$ is known,
the approximate $n$ distances can be computed in $O(km+kn)$
by the following algorithm:
\begin{equation} \label{wx}
w_x = V^T x, \quad
d_j^2 \approx \| w_x - w_j \|^2 ~~ \text{for $j=1,\ldots n$}
\end{equation}

\bsubsectionS{Our contributions}
Our main result is formulas that improve the
quality of the approximations in \reff{aiaj} and \reff{wx}.
Clearly, the approximation in \reff{aiaj}, and sometimes also
the approximation in \reff{wx},
can be improved by increasing $k$, 
the rank of the reduced dimension.
But this increases the computation cost and reduces the effectiveness
of working in a reduced dimension.
It also makes the interpretation of the data in the low dimension
harder.
For example, with $k=2$ the data can be visualized in a plane.
Increasing $k$ to $4$ creates a representation that is much harder to
visualize. 

Our main result is new formulas that improve the accuracy
in \reff{aiaj} and \reff{wx} without increasing $k$.
Specifically, we propose
the approximation formula~\reff{d3a}
as an alternative to \reff{aiaj},
and the approximation formula~\reff{d3x}
as an alternative to \reff{wx}:

\begin{align} \label{d3a}
& \text{distance}^2(a_i, a_j)  \approx \notag
\\
& ~~~~  \| w_i - w_j \|^2 + \|a_i\|^2 - \|w_i\|^2 + \|a_j\|^2 - \|w_j\|^2
\end{align}
\begin{align} \label{d3x}
  & \text{distance}^2(x, a_j) \approx \notag
  \\
  & ~~~~ \| w_x - w_j \|^2 + \|x\|^2 - \|w_x\|^2 + \|a_j\|^2 - \|w_j\|^2
  \\
  & \text{where~~} w_x = V^T x \notag
\end{align}
The following are some observations about the result:
\begin{compactitem}
\item
The formulas \reff{d3a} and \reff{d3x}
use additional information, the squared norms $\|a_i\|^2$ for each
column $a_i$ of $A$.
This information can be pre-computed during the PCA calculation,
without significant change to the running time of the PCA.
\item
  The complexity of using~\reff{d3a}
to compute the approximate squared distance between $a_i$ and $a_j$ 
is $O(k)$,
the same as the complexity of using~\reff{aiaj}
to approximate the squared distance between $a_i$ and $a_j$.
\item 
  The complexity of using~\reff{d3x}
  to compute the
  approximate squared distances
  between $x$ and all the columns of $A$ is $O(km + kn)$,
  the same as the complexity of using~\reff{wx}
  to compute the
  approximate squared distances
  between $x$ and all the columns of $A$.
\item
  The new approximations \reff{d3a} and \reff{d3x} are not always
  better than the old approximations \reff{aiaj} and \reff{wx}.
  But we claim that ``on the average'' the new  approximations are
  better.
  This follows from the derivation of these approximations
  using the Maximum Entropy Method and extensive
  evaluation on real datasets.
\end{compactitem}

\noindent
The technique that we use to derive the formulas
\reff{d3a} and \reff{d3x}
can also be used for other applications of the PCA besides
Euclidean distances.
We derive related formulas for accurate computation of the Rayleigh Quotient,
an important statistic that indicates the similarity of a vector
to a collection of vectors.

Another important contribution of the paper is the method in which the
approximations are derived.
We model the uncertainty in the PCA representation in terms of random variables 
with an unknown distribution.
We then use the Maximum Entropy Method to determine the most likely
distribution.
Expected values are then used as the improved estimates.
This approach appears to be novel.
We are not aware of any previous studies that apply similar approaches
to improve deterministic estimates.

\bsubsectionS{Paper organization}
The paper is organized as follows.
Section~\ref{secDR} formulates dimension reduction as an approximate
estimation of column vectors with unknown quantities.
A key idea is to model the unknown quantities as random
variables in an unknown probability distribution.

Section~\ref{secME} describes the Maximum Entropy Method,
a classical method of inferring
the most likely probability distribution from partial information about random
variables.
We apply the Maximum Entropy Method to the random variables
of Section~\ref{secDR} to derive the most likely probability distribution
of the PCA estimates.
A key theorem proved in this section characterizes the probability density
of the unknown quantities.

In Section~\ref{EVPCA}
we use the probability density of Section~\ref{secME} to compute
expected values of several expressions of PCA approximations.

In Section~\ref{secDistances}
we apply the results of Section~\ref{EVPCA} to compute
estimates to distances between vectors.
In Section~\ref{secRQ}
we derive maximum entropy estimates to Rayleigh quotients.

Section~\ref{secExperiments} describes extensive experimental results
evaluating our approximation formulas on real data.

\bsection{A probabilistic setting for PCA} \label{secDR}
As discussed in Section~\ref{secIntro}
the PCA approximation of the $m \times n$ matrix $A$ is
given by \Ereff{AVW}.
In this section we use a slightly different notation for the same
relation. We write the PCA approximation as:
\begin{equation} \label{AV1W1}
A \approx V_1 W_1
\end{equation}
Since $V_1$ has orthogonal columns, it is always possible to extend these
columns to an orthogonal basis of $\fatR^m$.
Let $V_2$ be such an extension then $V_1$ and $V_2$ are orthogonal complements.
They satisfy the following properties:
\begin{equation} \label{V1V2}
V_1^T V_1 = I, \quad V_2^T V_2 = I, \quad V_1 V_1^T + V_2 V_2^T = I
\end{equation}
Using both $V_1$ and $V_2$ there is an exact representation of $A$
that can be expressed as follows:
\begin{equation} \label{AV1W1V2W2}
  A  = V_1 W_1 + V_2 W_2,
  \quad
  a_i = V_1 w_1^i + V_2 w_2^i
\end{equation}
where $a_i$ is the $i$th column of $A$,
$w_1^i$ is the $i$th column of $W_1$,
and
$w_2^i$ is the $i$th column of $W_2$.
Suppose the PCA of $A$ is given as the matrices
$V_1$ and $W_1$. Without loss of generality
$V_2$ can be selected as any orthogonal complement of $V_1$.
This means that the only unknown quantities in \reff{AV1W1V2W2}
are the entries of the matrix $W_2$.
The special case of classical PCA is obtained by taking $W_2=0$.
Instead, we propose to view $W_2$ as a random matrix,
with entries that are random variables.
From  \reff{AV1W1V2W2} it follows that if $W_2$ is a random matrix
then $A$ is also a random matrix, and so are the columns $a_i$ and $w_2^i$.
\Ereff{rvA} identifies random variables with $\rv{~}$
as shown below:
\begin{equation} \label{rvA}
  \rv{A}  = V_1 W_1 + V_2 \rv{W_2},
  \quad
  \rv{a_i} = V_1 w_1^i + V_2 \rv{w_2^i}
\end{equation}
We note that some of the matrices in \reff{rvA} are 
too big to manipulate explicitly.
The size of $V_2$ is $ m \times m{-}k$,
and the size of $W_2$ is $ m{-}k \times n$.
A practical solution should not manipulate these matrices explicitly.

We proceed to show that
modeling the unknown $W_2$ as a random matrix 
has an advantage over setting it to be $0$.
Suppose the probability density of $W_2$ is \underline{known}.
Applying the expectation operator $\ex{}$ to both sides 
of the first equation in~\reff{rvA}
we get:
\[
\ex{\rv{A}}  = V_1 W_1 + V_2 \ex{\rv{W_2}}
\]
Thus, 
Taking $\ex{\rv{A}}$ as an improved estimate of $A$ we can expect an improved
result different from the classical result whenever $\ex{\rv{W_2}}$ is nonzero.
Similarly, using the orthogonality of $V_1,V_2$
it is easy derive the following relation from~\reff{rvA}:
$\rv{A^T}\rv{A}  =  W_1^T W_1 + \rv{W_2^T}\rv{W_2}$.
Taking expectations we see that:
\[
\ex{\rv{A^T}\rv{A}}  =  W_1^T W_1 + \ex{\rv{W_2^T}\rv{W_2}}
\]
Therefore, the improved estimate of $A^TA$ is different
from the classical estimate
whenever $\ex{\rv{W_2^T}\rv{W_2}} \neq 0$.
Observe that $\ex{\rv{W_2^T}\rv{W_2}} \neq \ex{\rv{W_2}}^T \ex{\rv{W_2}}$,
so that $\ex{\rv{W_2^T}\rv{W_2}}$ may be nonzero even if $\ex{\rv{W_2}}$ is 0.

In our case the probability density of $W_2$ is unknown.
We use the Maximum Entropy Method to compute the most likely probability
distribution under the assumption that the column norms of $A$ are known.
It is not surprising that under this probability density 
$\ex{\rv{W_2}} = 0$,
but we found it surprising that $\ex{\rv{W_2^T}\rv{W_2}} \neq 0$.

\bsection{The Maximum Entropy Method} \label{secME}

The Maximum Entropy Method is a standard technique for inferring
probability distributions.
When given constraints that the probability distribution must
satisfy, the Maximum Entropy Method asserts 
that the ``most likely distribution'' 
is the distribution with the largest entropy that
satisfies the constraints.
See, e.g.,~\cite{Jaynes82,papoulis,wiki:MaxEntropy}.
In this paper we use the following special case
described in Chapter 14 in Papoulis~\cite{papoulis}.

\smallskip
\noindent
\textbf{Theorem~1:}
~
Let $x = (x_1, \ldots, x_n)^T$ be a random vector, where the coordinates
$x_i$ are $n$ random variables.
Let $R = \ex{x x^T}$ be the correlation matrix associated with $x$.
Suppose $R$ is known.
Let $\Delta$ be the determinant of $R$ and assume $ \Delta \neq 0$.
Then according to the Maximum Entropy Method the probability density
$f(x)$ and the entropy $H(x)$ are given by:
\begin{equation} \label{fH}
\begin{aligned}
& f(x) = \frac{1}{\sqrt{(2 \pi)^n \Delta }} e^{-\frac{1}{2}x^t R^{-1} x}
\\
& H(x) = \ln \sqrt{(2 \pi e)^n \Delta}
\end{aligned}
\end{equation}
\noindent
See~\cite{papoulis} for the proof.

\smallskip
\noindent
As stated in~\reff{fH}
the entropy of $x$ is determined by $\Delta$,
the determinant of the correlation matrix $R$.
If $R$ is only partially known,
it can be determined by the Maximum Entropy Method by
maximizing the determinant $\Delta$ over the unknown quantities.
We use this technique to derive the following theorem which is
our main technical result:

\smallskip
\noindent
\textbf{Theorem~2:}
~
Let $W = (w_1, \ldots, w_n)$ be a random matrix
of dimensions $k \times n$.
Suppose $z_i = \ex{\|w_i\|^2}$ is known for $i=1,\ldots,n$,
but nothing else is known about the probability density of $W$.
Then according to the Maximum Entropy Method:
\begin{compactdesc}
\item[1.]
  All entries of the matrix $W$ have 0 mean:
  \[
  \ex{w_{ij}} = 0, \quad \text{for}~ i=1,\ldots, n, ~ j=1,\ldots,k
  \] 
\item[2.]
  The random variable $w_{i_1,j_1}$
  is independent of the random variable $w_{i_2,j_2}$
  unless $i_1 = i_2$ and $j_1 = j_2$.
\item[3.] The expected value of $\|w_{ij}\|^2$ is given by:
\[
\ex{\|w_{ij}\|^2} = \frac{z_i}{k}
\]
\item[4.]
  The probability density of $W$ is given by:
  \[
  \begin{aligned}
    & f(W) = \frac{1}{\sqrt{(2 \pi)^{kn} \Delta }} e^{s(W)}
    \\
    & \text{where:}
    \\
    & \Delta = \frac{\prod_{i=1}^n z_i^k}{k^{kn}},
    \quad
    s(W) = -\frac{k}{2} \sum_{i,j} \frac{w_{i,j}^2}{z_i}
  \end{aligned}
\]
\end{compactdesc}

\smallskip
\noindent
\textbf{Proof:}
~
Let $q$ be the vector of $kn$ random variables, created by concatenating
all the columns of $W$:
\[
q = \begin{pmatrix}
  w_{11}, \ldots, w_{1k},
  \ldots, 
  w_{n1}, \ldots, w_{nk}
\end{pmatrix}
^T
\]
Let $R$ be the correlation matrix of $q$:
$R = \ex{q q^T}$.
Observe that the matrix $R$ is $nk \times nk$.
The value of $R$ at row $I$ and column $J$ is:
$R_{IJ} = \ex{w_{i_1,j_1} w_{i_2,j_2}}$ for some $i_1,j_1,i_2,j_2$.
Define:  $\nu_{ij} = \ex{w_{ij}^2}$.
Then from the definition of $z_i$:
\begin{equation} \label{sumnu}
  z_i = \ex{\|w_i\|^2}
  % = \ex{\sum_j w_{ij}^2}
  = \sum_j \ex{w_{ij}^2} = \sum_j \nu_{ij}
\end{equation}
The following information is known about the diagonal elements $R_{II}$,
where the location $I$ in $q$ correspond to the location $i,j$ in $W$:
\[
R_{II} = \ex{(w_{ij}}^2) = \nu_{ij}.
\]
From Theorem~1 it follows that the maximum entropy of $q$ is obtained
by maximizing $\Delta$, the determinant of $R$,
under the constraints~\reff{sumnu}. 
The proof of the theorem follows from this maximization.

According to the Hadamard determinant inequality
(see, e.g., \cite{Rozanski17}), 
$\Delta \leq \prod_I R_{II}$.
Since 
$\Delta = \prod_I R_{II}$ if all off diagonal elements are 0,
it follows that there is a maximum where $\ex{w_{i_1,j_1} w_{i_2,j_2}} = 0$
unless $i_1 {=} i_2$ and $j_1 {=} j_2$.
This proves parts 1 and 2 of the theorem.

To prove part 3 observe that from parts 1,2 it follows that
\[
\Delta = \prod_I R_{II} = \prod_{i,j} \nu_{ij}
= \prod_i (\prod_j \nu_{ij})
\]
Therefore, for each $i$ we need to maximize  $\prod_{i,j} \nu_{ij}$
subject to the constraint that $\sum_{j=1}^k \nu_{ij} = z_i$.
It can be easily shown (for example using the method of Lagrange multipliers)
that the maximizing solution is $\nu_{ij} = \frac{z_i }{k}$.
This proves part 3 of the theorem.
Part 4 of the theorem follows from Theorem 1 by observing that:
\begin{compactitem}
\item
  From part 3:
  \[
  \Delta = \prod_{i=1}^n \prod_{j=1}^k (z_i/k)
  = \frac{\prod_{i=1}^n z_i^k}{k^{kn}}
  \]
\item
  Since $R$ is diagonal:
  \[
  q^T R^{-1} q = \sum_I \frac{q_I^2}{R_{II}} =
    \sum_{i,j} \frac{w_{i,j}^2}{z_i/k}
  \]
\end{compactitem}
This completes the proof of Theorem~2.
\hfill $\blacksquare$

\bsection{Expected values of PCA approximations} \label{EVPCA}
In \Ereff{rvA}
the PCA is expressed as a relation between random variables.
We apply Theorem~2 to the matrix $W_2$ and determine its most likely
probability density.
The expected values of various estimates can then be computed in closed form.
The matrix $W_2$ is of size $m{-}k \times n$,
and its $i$th column, $w_2^i$, is of size $m{-}k$.
The value of $z_i$ in Theorem 2 can be computed
as follows:
\begin{equation} \label{zi}
  z_i = \ex{\|\rv{w_2^i}\|^2} = \|a_i\|^2 - \|w_1^i\|^2, \quad i=1,\ldots,n
\end{equation}
Applying Theorem~2 this gives the following expected values
of expressions related to $w_2^i$, the $i$th column of $W_2$:
\begin{equation} \label{w2}
  \begin{aligned}
    & \ex{ \rv{w_2^i} } = 0
    , \quad \ex{\rv{\|w_2^i\|^2}} = z_i
    \\
    & \ex{\rv{(w_2^i)^T}\rv{w_2^j}} = 0
      ,\quad \ex{\rv{w_2^i}\rv{(w_2^i)^T}} = \frac{z_i}{m-k} I
      \\
      & \text{where $i {\neq} j$, 
        and $I$ is the $m{-}k \times m{-}k$ identity matrix.}
  \end{aligned}
\end{equation}
From~\reff{w2} we get the following
expected values related to the entire matrix $W_2$:
\begin{equation} \label{W2}
\begin{aligned}
& \ex{\rv{W_2}} = 0
\\
& \ex{\rv{W_2^T}\rv{W_2}} = 
\begin{pmatrix}
z_1 \\ &  z_2 \\ & & \ddots & \\ &  & & z_n
\end{pmatrix}
\\
& \ex{\rv{W_2}\rv{W_2^T}} = \frac{\sum_{i=1}^n z_i}{m-k} I
= \delta I
\\
& \text{where $\delta = {\sum_{i=1}^n z_i} / {(m-k)}$}
\end{aligned}
\end{equation}

The corresponding formulas for $\rv{A} = V_1 W_1 + V_2 \rv{W_2}$
(as in~\reff{rvA}) are:
\begin{equation} \label{AA}
\begin{aligned}
& \ex{\rv{A}} = V_1 W_1
\\
& \ex{\rv{A^T}\rv{A}} = W_1^T W_1 + \text{Diag}(z_1, \ldots z_n)
\\
& \ex{\rv{A}\rv{A^T}} = V_1 W_1 W_1^T V_1^T + \delta (I - V_1 V_1^T)
\end{aligned}
\end{equation}
The first equation follows from the first formula in~\reff{W2}.
The second equation
follows by applying expectations to the identity:
$\rv{A^T}\rv{A}= W_1^T W_1 + \rv{W_2^T}\rv{W_2}$.
The third equation
follows by applying expectations to the identity:
\[
\begin{aligned}
\rv{A}\rv{A^T} = &
V_1 W_1 W_1^T V_1^T
+ V_2 \rv{W_2}\rv{W_2^T} V_2^T
\\
& + V_1 W_1 \rv{W_2^T} V_2^T + V_2 \rv{W_2} W_1^T V_1^T
\\
\end{aligned}
\]
Taking expectations the last two terms disappear.
The final result is obtained by applying~\reff{W2}
to second expression,
and using~\reff{V1V2}
to replace $V_2V_2^T$ with $I-V_1V_1^T$.

\bsection{Computing distances with PCA} \label{secDistances}
In this section we assume being given the matrix $A$
with pre-computed PCA expressed as: $A \approx V_1 W_1$.
In addition to the PCA
we assume that the column norms $\|a_i\|$ are known for
all the columns of $A$.
Two cases are analyzed.
In the first case the goal is to compute distances between
columns of $A$.
In the second case the goal is to compute distances between
a vector $x$ unrelated to $A$ and columns of $A$.
In each case we describe three formulas.
The first formula that we denote by $\dc$ is the classical formula.
It does not use the additional information of column norms.
The second formula that we denote by $\de$ is obtained from the
Maximum Entropy Method.
It requires the additional information of column norms.
Since $\de$ works much better than $\dc$ one may suspect that
the reason might be additional information of column norms.
We use this information to derive another distance formula,
as a tight lower bound to the true distance that also requires
the additional information of column norms.
We denote this third distance formula by $\dl$.
Our experimental results show that typically
$\dl$ is much better than $\dc$, and 
$\de$ is much better than $\dl$.

\bsubsection{Distances between columns of $A$} \label{secdaa}
We consider approximating the distance between $a_i$ and $a_j$, 
two columns of $A$.
Their PCA representation is:
\begin{equation} \label{aiajPCA}
  a_i \approx V_1 w_1^i, \quad a_j \approx V_1 w_1^j
\end{equation}
As discussed in Section~\ref{secIntro} 
the classical approximation formula for the squared distance between them is:
\begin{equation} \label{dci}
\text{distance}^2(a_i,a_j) \approx
\dc(a_i,a_j) = \|w_1^i - w_1^j\|^2
\end{equation}

\smallskip
\noindent
\textbf{Theorem~3:}
~
Let $a_i$ and $a_j$ be two columns of $A$ with PCA representation as 
shown in~\reff{aiajPCA}.
The estimate of the squared distance between them according to
the Maximum Entropy Method is:
\[
\begin{aligned}
& \text{distance}^2(a_i,a_j) \approx
\\
& \qquad \dc(a_i,a_j) + \|a_i\|^2 - \|w_1^i\|^2 + \|a_j\|^2 - \|w_1^j\|^2
\end{aligned}
\]
\noindent
\textbf{Proof:}
~
The random variable representation in~\reff{rvA} gives:
\[
  \rv{a_i} = V_1 w_1^i + V_2 \rv{w_2^i}, \quad
  \rv{a_j} = V_1 w_1^j + V_2 \rv{w_2^j}
\]
Computing the squared Euclidean distance between them as a random variable
gives: 
\[
\begin{aligned}
  &  \|\rv{a_i}-\rv{a_j}\|^2 =
  \| V_1 (w_1^i - w_1^j) + V_2 (\rv{w_2^i} - \rv{w_2^j}) \|^2
\\
& = \| w_1^i - w_1^j \|^2 + \| \rv{w_2^i} - \rv{w_2^j} \|^2 
\\
& = \| w_1^i - w_1^j \|^2 + \| \rv{w_2^i} \|^2 + \| \rv{w_2^j} \|^2 
- 2 (\rv{w_2^i})^T \rv{w_2^j}
\end{aligned}
\]
Going to expectations and using \Ereff{w2}
we see that the expected value of the right most term is 0,
and the values of the middle two terms are $z_i$, $z_j$.
This gives:
\begin{equation} \label{de}
\begin{aligned}
& \de(a_i,a_j) = \dc(a_i,a_j) +  z_i + z_j
\end{aligned}
\end{equation}
The theorem now follows from~\reff{zi}.
\hfill $\blacksquare$

\smallskip
\noindent
\textbf{Theorem~4:}
~
Define $\dl$ as follows:
\[
\begin{aligned}
& \dl(a_i,a_j) = \dc(a_i,a_j) + z_i + z_j - 2 \sqrt{z_i z_j}
\\
& \text{where $z_i = \| a_i\|^2 - \|w_1^i\|^2$}
\end{aligned}
\]
Then:
\[
\dc(a_i,a_j) \leq \dl(a_i,a_j) \leq \text{distance}^2(a_i,a_j)
\]
\noindent
\textbf{Proof:}
~
The following relations hold:
\[
\begin{aligned}
a.~& \text{distance}^2(a_i,a_j) = \|w_1^i - w_1^j\|^2 + \|w_2^i - w_2^j\|^2
\\
b.~&\|w_2^i - w_2^j\|^2 \geq (\|w_2^i\| - \|w_2^j\|)^2
= z_i + z_j - 2 \sqrt{z_i z_j}
\\
c.~& \dc = \|w_1^i - w_1^j\|^2
\end{aligned}
\]
Relation $a$ follows from~\reff{AV1W1V2W2}.
Relation $b$ follows from the triangle inequality.
Relation $c$ is the definition of $\dc$.
Combining relations $b$ and $c$ gives the left inequality in the theorem.
Combining relations $a$ and $b$ gives the right inequality in the theorem.
\hfill $\blacksquare$

\noindent
This shows that $\dl$ is a lower bound on the true distance.
The bound is tight since there is an equality in $b$ when
the angle between $w_2^i$ and $w_2^j$ is 0.

\smallskip
\noindent
In summary, we describe 3 formulas for estimating distances between
matrix columns using PCA data: 
\[
\begin{aligned}
&\dc(a_i,a_j) = \|w_1^i - w_1^j\|^2
\\
&\dl(a_i,a_j) = \dc(a_i,a_j) + z_i + z_j - 2 \sqrt{z_i z_j}
\\
&\de(a_i,a_j) = \dc(a_i,a_j) +  z_i + z_j
\end{aligned}
\]
\\
In these formulas $w_1^i$ is the representation of column $a_i$ in PCA space,
and $z_i = \|a_i\|^2 - \|w_i\|^2$.
Column norms are used
by $\de$ and $\dl$. Both $\dc$ and $\dl$ are lower bounds on
the true distance, and $\dl$ is guaranteed to be better than $\dc$.
The promise of $\de$ is that it was derived from the best probability
distribution according to the Maximum Entropy Method.
As we show in the experimental section its accuracy is significantly better
than the accuracy of $\dl$ and $\dc$.

\bsubsection{Distances between an arbitrary vector and columns of $A$}
\label{secdax}
Let $x$ be an arbitrary ($m$ dimensional) vector.
Our goal is to approximate efficiently and accurately distances
between $x$ and the columns of $A$.
As in Section~\ref{secdaa} we assume the availability of 
the PCA of $A$, as well as the norms of $A$ columns.
We begin by defining the vectors $w_1^x$ and $w_2^x$ as analogous 
to $w_1^i$ and $w_2^i$ for a column of $A$:
\begin{equation} \label{w12x}
w_1^x = V_1^T x, \quad w_2^x = V_2^T x
\end{equation}
With this definition most of the analysis in Section~\ref{secdaa}
applies to case as well.
The only difference in the analysis is that $w_2^x$ can be explicitly
calculated, and therefore it is not a random variable.
Still, the three distance formulas from Section~\ref{secdaa}
can be used in this case as well. 
As in~\reff{dci} the classical error estimate is given below:
\begin{equation} \label{dcx}
\text{distance}^2(x,a_j) \approx
\dc(x,a_j) = \|w_1^x - w_1^j\|^2
\end{equation}
This approximation can only be accurate when $x$ projection on $V2$ is small.

\smallskip
\noindent
\textbf{Theorem~5:}
~
Define $\dl$ as follows:
\[
\begin{aligned}
& \dl(x,a_j) = \dc(x,a_j) + z_x + z_j - 2 \sqrt{z_x z_j}
\\
& \text{where $z_x = \|x\|^2 - \|w_1^x\|^2$, $z_j = \| a_j\|^2 - \|w_1^j\|^2$}
\end{aligned}
\]
Then:
\[
\dc(x,a_j) \leq \dl(x,a_j) \leq \text{distance}^2(x,a_j)
\]
\noindent
\textbf{Proof:}
This is identical to the proof of Theorem~4.
\hfill $\blacksquare$

\smallskip
\noindent
\textbf{Theorem~6:}
~
The estimate of the squared distance between $x$ and a column of $A$
 according to the Maximum Entropy Method is:
\[
\begin{aligned}
& \text{distance}^2(x,a_j) \approx
\\
& \qquad \dc(x,a_j) + \|x\|^2 - \|w_1^x\|^2 + \|a_j\|^2 - \|w_1^j\|^2
\end{aligned}
\]
\noindent
\textbf{Proof:}
~
Both $x$ and the random variable $a_j$ can be described in terms of their
projections on $V_1,V_2$. 
\[
  x = V_1 w_1^x + V_2 w_2^x, \quad
  \rv{a_j} = V_1 w_1^j + V_2 \rv{w_2^j}
\]
Calculating the squared norm of their difference:
\[
\begin{aligned}
  &  \|x -\rv{a_j}\|^2
  \\
  & ~~ =
  \|x\|^2 + \|w_1^j\|^2 + \|\rv{w_2^j}\|^2
  - 2 (w_1^x)^T w_1^j -2 (w_2^x)^T \rv{w_2^j}
\\
& ~~ = 
\| w_1^x - w_1^j \|^2
- \| w_1^x\|^2
+ \| \rv{w_2^j}\|^2
+ \|x\|^2
- 2(w_2^x)^T \rv{w_2^j}
\end{aligned}
\]
Going to expectations and using \Ereff{w2}
we get:
\[
\begin{aligned}
& \de(x,a_j) = \dc(x,a_j) +  \|x\|^2 - \|w_1^x\|^2 + z_j
\end{aligned}
\]
The theorem now follows from~\reff{zi}.
\hfill $\blacksquare$

\bsection{Rayleigh Quotients} \label{secRQ}

The Rayleigh Quotients (e.g., \cite{gv4}) is given by the following formula:
\begin{equation} \label{RQ}
r(v) = \frac{v^T B v}{\|v\|^2}
\end{equation}
For a given matrix $A$
we are interested in the two special cases
of $B=AA^T$, and $B=A^TA$.
In the first case the Rayleigh Quotient
gives the sum of squared correlation between $v$ and the columns of $A$.
In the second case it gives the sum of squared correlation between $v$ and the
rows of $A$. 
Intuitively, 
the Rayleigh quotients measure the likelihood 
of the direction of $v$ among the columns/rows of the matrix $A$.

The challenge we address here is how to estimate these 
Rayleigh Quotients when given the PCA of $A$ instead of $A$ itself.
The classical solution is to replace $A$ with its PCA representation,
as given by~\reff{AV1W1}.
As in the case of distances the Maximum Entropy Method gives 
an improved solution.

\bsubsection{Column space Rayleigh quotient} \label{seccs}
In this section we consider the case in which $B=AA^T$ 
in~\reff{RQ}.
For a vector $x \in \fR{m}$ the exact expression we wish to approximate
is:
\[
r(x) = \frac{x^T A A^T x}{\|x\|^2}
\]
When $A$ is approximated as in~\reff{AV1W1} we have:
\begin{equation} \label{rc1}
\begin{aligned}
& \rc(x) = 
  \frac{x^T V_1 W_1 W_1^T V_1^T x }{\|x\|^2}
  = \frac{\|W_1^T w_1^x\|^2 }{\|x\|^2}
\\
& \text{where $w_1^x = V_1^T x$}.
\end{aligned}
\end{equation}
For the derivation of the Maximum Entropy solution we use the representation
of $A$ as a random matrix in~\reff{rvA}.
\[
\rv{r(x)} = \frac{x^T \rv{A}\rv{A^T}x}{\|x\|^2} 
\]
Taking expectations of both side and using the result in~\Ereff{AA}
we get:
\[
\begin{aligned}
& \re =
\frac{x^T V_1 W_1 W_1^T V_1^T x + \delta (\|x\|^2 - x^T V_1 V_1^T x)}{\|x\|^2}
\\
&
= \frac{x^T V_1 (W_1 W_1^T - \delta I) V_1^T x} {\|x\|^2} + \delta
\\
&
= \frac{\|W_1^T w_1^x\|^2}{\|x\|^2} 
+ \delta (1 - \frac{\|w_1^x\|^2}{\|x\|^2} )
=
\rc
+ \delta (1 - \frac{\|w_1^x\|^2}{\|x\|^2} )
\end{aligned}
\]

\bsubsection{Row space Rayleigh quotient} \label{secrs}
In this section we consider the case in which $B=A^TA$ 
in~\reff{RQ}.
For a vector $y \in \fR{n}$ the exact expression we wish to approximate
is:
\[
r(y) = \frac{y^T A^T A y}{\|y\|^2}
\]
When $A$ is approximated as in~\reff{AV1W1} we have:

\begin{equation} \label{rr1}
\rc(y) =
 \frac{y^T W_1^T W_1 y }{\|y\|^2}
= \frac{\|W_1 y\|^2}{\|y\|^2}
\end{equation}

For the derivation of the Maximum Entropy solution we use the representation
of $A$ as a random matrix in~\reff{rvA}.
\[
\rv{r(y)} = \frac{y^T \rv{A^T}\rv{A}y}{\|y\|^2} 
\]
Taking expectations of both side and using the result in~\Ereff{AA}
we get:
\[
\begin{aligned}
\re(y) & = \frac{y^T 
( W_1^T W_1 + \text{Diag}(z_1, \ldots z_n))
y}{\|y\|^2} 
\\
&=
\rc(y) + \frac{y^T \text{Diag}(z_1, \ldots z_n) y}{\|y\|^2}
\\
&=
\rc(y) + \frac{\sum_{i=1}^n z_i (y(i))^2}{\|y\|^2}
\end{aligned}
\]
In summary we have the following formulas:
\begin{equation} \label{RQf}
  \begin{aligned}
    & \text{column space} &&
    \rc(x)  = \frac{\|W_1^T w_1^x\|^2 }{\|x\|^2}
    \\
    & \text{column space} &&
    \re(x) =
    \rc + \delta (1 - \frac{\|w_1^x\|^2}{\|x\|^2} )
    \\
    & \text{row space} &&
    \rc(y) = \frac{\|W_1 y\|^2}{\|y\|^2}
    \\
    & \text{row space} &&
    \re = \rc +  \frac{ \sum_{i=1}^n z_i (y(i))^2}{\|y\|^2}
  \end{aligned}
\end{equation}

\bsection{Experimental results} \label{secExperiments}

\newcommand{\wipe}[1]{}

We ran many experiments on various real datasets from the
UC Irvine repository.
In all cases the formulas derived using the Maximum Entropy Method
produced better results than the classical formulas.
The improvements were very significant on most datasets.
The worst case was for the ``wdbc'' dataset, shown later.
Experiments on three datasets are described in detail.
They include the ``Ionosphere'' (size $33 \times 351$),
the ``wdbc'' (size $30 \times 569$),
and the ``YearPredictionMDS'' (size $90 \times 515,345$).

To experiment with column distances we measured the distances between
all pairs of columns of the data matrix.
In each case we compute the difference (in absolute value)
between the computed distance and the true distance.
This is done for various $k$ values. 
For the ``YearPredictionMSD'' dataset, because of the large number of observations 
we selected 50 columns at random,
and computed the distances between all pairs in the selection.
To measure the distances between an arbitrary vector $x$ and columns of $A$,
the vector $x$ was drawn from Gaussian distribution with mean=0 and variance=1.

% tables

 %../dataset/ionosphere_nolabel.csv
\begin{table}[h] 
\caption{Distance(Ionosphere)}
\label{tab:Distanceionosphere} 
 
  \begin{adjustbox}{width=\columnwidth} 
  \begin{tabular}{|c|c|c|c|c|c|} 
  \hline 
  
  \multicolumn{2}{|c|}{~} & 
  \multicolumn{2}{|c|}{$x$ and columns of $A$} & 
  \multicolumn{2}{|c|}{column distances} \\ 
  \cline{3-6} 

  \multicolumn{2}{|c|}{~} &mean & std & mean & std \\
  \hline

\multirow{3}{*}{k= 1} 
&$\lvert \dc-d \rvert$ & 
\multicolumn{1}{|c|}{ 3.905E+01} &  \multicolumn{1}{|c|}{ 1.218E+01} &\multicolumn{1}{|c|}{ 1.382E+01} &  \multicolumn{1}{|c|}{ 1.075E+01} \\ 
&$\lvert  \dl-d \rvert$ & 
\multicolumn{1}{|c|}{ 2.529E+01} &  \multicolumn{1}{|c|}{ 1.743E+01} &\multicolumn{1}{|c|}{ 9.650E+00} &  \multicolumn{1}{|c|}{ 1.021E+01} \\ 
&$\lvert \de-d \rvert$ & 
\multicolumn{1}{|c|}{ \textbf{ 3.645E+00}} &  \multicolumn{1}{|c|}{ \textbf{ 3.976E+00}} &\multicolumn{1}{|c|}{ \textbf{ 2.161E+00}} &  \multicolumn{1}{|c|}{ \textbf{ 3.585E+00}} \\ 
\hline 

% \multirow{3}{*}{k= 2} 
% &$\lvert \dc-d \rvert$ & 
% \multicolumn{1}{|c|}{ 3.699E+01} &  \multicolumn{1}{|c|}{ 1.199E+01} &\multicolumn{1}{|c|}{ 1.164E+01} &  \multicolumn{1}{|c|}{ 1.012E+01} \\ 
% &$\lvert \dl-d \rvert$ & 
% \multicolumn{1}{|c|}{ 2.221E+01} &  \multicolumn{1}{|c|}{ 1.606E+01} &\multicolumn{1}{|c|}{ 8.017E+00} &  \multicolumn{1}{|c|}{ 8.942E+00} \\ 
% &$\lvert \de-d \rvert$ & 
% \multicolumn{1}{|c|}{ \textbf{ 3.222E+00}} &  \multicolumn{1}{|c|}{ \textbf{ 3.629E+00}} &\multicolumn{1}{|c|}{ \textbf{ 1.773E+00}} &  \multicolumn{1}{|c|}{ \textbf{ 2.687E+00}} \\ 
% \hline 

\multirow{3}{*}{k= 3} 
&$\lvert \dc-d \rvert$ & 
\multicolumn{1}{|c|}{ 3.474E+01} &  \multicolumn{1}{|c|}{ 1.096E+01} &\multicolumn{1}{|c|}{ 9.384E+00} &  \multicolumn{1}{|c|}{ 9.432E+00} \\ 
&$\lvert \dl-d \rvert$ & 
\multicolumn{1}{|c|}{ 1.870E+01} &  \multicolumn{1}{|c|}{ 1.537E+01} &\multicolumn{1}{|c|}{ 5.856E+00} &  \multicolumn{1}{|c|}{ 7.551E+00} \\ 
&$\lvert \de-d \rvert$ & 
\multicolumn{1}{|c|}{ \textbf{ 2.716E+00}} &  \multicolumn{1}{|c|}{ \textbf{ 3.327E+00}} &\multicolumn{1}{|c|}{ \textbf{ 1.171E+00}} &  \multicolumn{1}{|c|}{ \textbf{ 1.967E+00}} \\ 
\hline 

% \multirow{3}{*}{k= 4} 
% &$\lvert \dc-d \rvert$ & 
% \multicolumn{1}{|c|}{ 3.307E+01} &  \multicolumn{1}{|c|}{ 1.056E+01} &\multicolumn{1}{|c|}{ 8.084E+00} &  \multicolumn{1}{|c|}{ 8.981E+00} \\ 
% &$\lvert \dl-d \rvert$ & 
% \multicolumn{1}{|c|}{ 1.634E+01} &  \multicolumn{1}{|c|}{ 1.485E+01} &\multicolumn{1}{|c|}{ 4.621E+00} &  \multicolumn{1}{|c|}{ 6.753E+00} \\ 
% &$\lvert \de-d \rvert$ & 
% \multicolumn{1}{|c|}{ \textbf{ 2.458E+00}} &  \multicolumn{1}{|c|}{ \textbf{ 3.220E+00}} &\multicolumn{1}{|c|}{ \textbf{ 8.497E-01}} &  \multicolumn{1}{|c|}{ \textbf{ 1.680E+00}} \\ 
% \hline 

\multirow{3}{*}{k= 5} 
&$\lvert \dc-d \rvert$ & 
\multicolumn{1}{|c|}{ 3.169E+01} &  \multicolumn{1}{|c|}{ 9.788E+00} &\multicolumn{1}{|c|}{ 7.182E+00} &  \multicolumn{1}{|c|}{ 7.927E+00} \\ 
&$\lvert \dl-d \rvert$ & 
\multicolumn{1}{|c|}{ 1.523E+01} &  \multicolumn{1}{|c|}{ 1.375E+01} &\multicolumn{1}{|c|}{ 4.134E+00} &  \multicolumn{1}{|c|}{ 5.952E+00} \\ 
&$\lvert \de-d \rvert$ & 
\multicolumn{1}{|c|}{ \textbf{ 2.351E+00}} &  \multicolumn{1}{|c|}{ \textbf{ 3.027E+00}} &\multicolumn{1}{|c|}{ \textbf{ 7.608E-01}} &  \multicolumn{1}{|c|}{ \textbf{ 1.467E+00}} \\ 
\hline 

\multirow{3}{*}{k= 10} 
&$\lvert \dc-d \rvert$ & 
\multicolumn{1}{|c|}{ 2.496E+01} &  \multicolumn{1}{|c|}{ 7.874E+00} &\multicolumn{1}{|c|}{ 4.391E+00} &  \multicolumn{1}{|c|}{ 4.993E+00} \\ 
&$\lvert \dl-d \rvert$ & 
\multicolumn{1}{|c|}{ 1.054E+01} &  \multicolumn{1}{|c|}{ 9.897E+00} &\multicolumn{1}{|c|}{ 2.483E+00} &  \multicolumn{1}{|c|}{ 3.720E+00} \\ 
&$\lvert \de-d \rvert$ & 
\multicolumn{1}{|c|}{ \textbf{ 1.789E+00}} &  \multicolumn{1}{|c|}{ \textbf{ 2.383E+00}} &\multicolumn{1}{|c|}{ \textbf{ 4.655E-01}} &  \multicolumn{1}{|c|}{ \textbf{ 9.251E-01}} \\ 
\hline 

% \multirow{3}{*}{k= 15} 
% &$\lvert \dc-d \rvert$ & 
% \multicolumn{1}{|c|}{ 1.975E+01} &  \multicolumn{1}{|c|}{ 6.950E+00} &\multicolumn{1}{|c|}{ 2.678E+00} &  \multicolumn{1}{|c|}{ 3.063E+00} \\ 
% &$\lvert \dl-d \rvert$ & 
% \multicolumn{1}{|c|}{ 7.373E+00} &  \multicolumn{1}{|c|}{ 7.077E+00} &\multicolumn{1}{|c|}{ 1.512E+00} &  \multicolumn{1}{|c|}{ 2.295E+00} \\ 
% &$\lvert \de-d \rvert$ & 
% \multicolumn{1}{|c|}{ \textbf{ 1.403E+00}} &  \multicolumn{1}{|c|}{ \textbf{ 1.863E+00}} &\multicolumn{1}{|c|}{ \textbf{ 3.072E-01}} &  \multicolumn{1}{|c|}{ \textbf{ 6.158E-01}} \\ 
% \hline 

% \multirow{3}{*}{k= 20} 
% &$\lvert \dc-d \rvert$ & 
% \multicolumn{1}{|c|}{ 1.368E+01} &  \multicolumn{1}{|c|}{ 5.403E+00} &\multicolumn{1}{|c|}{ 1.522E+00} &  \multicolumn{1}{|c|}{ 1.718E+00} \\ 
% &$\lvert \dl-d \rvert$ & 
% \multicolumn{1}{|c|}{ 4.668E+00} &  \multicolumn{1}{|c|}{ 4.560E+00} &\multicolumn{1}{|c|}{ 8.702E-01} &  \multicolumn{1}{|c|}{ 1.319E+00} \\ 
% &$\lvert \de-d \rvert$ & 
% \multicolumn{1}{|c|}{ \textbf{ 1.054E+00}} &  \multicolumn{1}{|c|}{ \textbf{ 1.387E+00}} &\multicolumn{1}{|c|}{ \textbf{ 2.043E-01}} &  \multicolumn{1}{|c|}{ \textbf{ 4.013E-01}} \\ 
% \hline 

  \end{tabular}
  \end{adjustbox}
\end{table}

 %../dataset/YearPredictionMSD_nolabel.csv
\begin{table}[h] 
\caption{Distance(YearPredictionMSD)}
\label{tab:DistanceYearPredictionMSD} 
 
  \begin{adjustbox}{width=\columnwidth} 
  \begin{tabular}{|c|c|c|c|c|c|} 
  \hline 
  
  \multicolumn{2}{|c|}{~} & 
  \multicolumn{2}{|c|}{$x$ and columns of $A$} & 
  \multicolumn{2}{|c|}{column distances} \\ 
  \cline{3-6} 

  \multicolumn{2}{|c|}{~} &mean & std & mean & std \\
  \hline

\multirow{3}{*}{k= 20} 
&$\lvert \dc-d \rvert$ & 
\multicolumn{1}{|c|}{ 4.453E+05} &  \multicolumn{1}{|c|}{ 4.606E+05} &\multicolumn{1}{|c|}{ 8.808E+05} &  \multicolumn{1}{|c|}{ 6.820E+05} \\ 
&$\lvert \dl-d \rvert$ & 
\multicolumn{1}{|c|}{ 9.876E+03} &  \multicolumn{1}{|c|}{ 5.038E+03} &\multicolumn{1}{|c|}{ 7.094E+05} &  \multicolumn{1}{|c|}{ 5.553E+05} \\ 
&$\lvert \de-d \rvert$ & 
\multicolumn{1}{|c|}{ \textbf{ 9.739E+02}} &  \multicolumn{1}{|c|}{ \textbf{ 9.650E+02}} &\multicolumn{1}{|c|}{ \textbf{ 1.226E+05}} &  \multicolumn{1}{|c|}{ \textbf{ 2.111E+05}} \\ 
\hline

\multirow{3}{*}{k= 40} 
&$\lvert \dc-d \rvert$ & 
\multicolumn{1}{|c|}{ 1.034E+05} &  \multicolumn{1}{|c|}{ 9.739E+04} &\multicolumn{1}{|c|}{ 2.047E+05} &  \multicolumn{1}{|c|}{ 1.463E+05} \\ 
&$\lvert \dl-d \rvert$ & 
\multicolumn{1}{|c|}{ 3.986E+03} &  \multicolumn{1}{|c|}{ 1.913E+03} &\multicolumn{1}{|c|}{ 1.701E+05} &  \multicolumn{1}{|c|}{ 1.236E+05} \\ 
&$\lvert \de-d \rvert$ & 
\multicolumn{1}{|c|}{ \textbf{ 4.451E+02}} &  \multicolumn{1}{|c|}{ \textbf{ 4.330E+02}} &\multicolumn{1}{|c|}{ \textbf{ 2.998E+04}} &  \multicolumn{1}{|c|}{ \textbf{ 4.735E+04}} \\ 
\hline 

% \multirow{3}{*}{k= 45} 
% &$\lvert \dc-d \rvert$ & 
% \multicolumn{1}{|c|}{ 7.282E+04} &  \multicolumn{1}{|c|}{ 7.656E+04} &\multicolumn{1}{|c|}{ 1.443E+05} &  \multicolumn{1}{|c|}{ 1.149E+05} \\ 
% &$\lvert \dl-d \rvert$ & 
% \multicolumn{1}{|c|}{ 3.274E+03} &  \multicolumn{1}{|c|}{ 1.696E+03} &\multicolumn{1}{|c|}{ 1.183E+05} &  \multicolumn{1}{|c|}{ 9.033E+04} \\ 
% &$\lvert \de-d \rvert$ & 
% \multicolumn{1}{|c|}{ \textbf{ 3.981E+02}} &  \multicolumn{1}{|c|}{ \textbf{ 3.807E+02}} &\multicolumn{1}{|c|}{ \textbf{ 2.228E+04}} &  \multicolumn{1}{|c|}{ \textbf{ 3.647E+04}} \\ 
% \hline 

\multirow{3}{*}{k= 50} 
&$\lvert \dc-d \rvert$ & 
\multicolumn{1}{|c|}{ 4.914E+04} &  \multicolumn{1}{|c|}{ 5.302E+04} &\multicolumn{1}{|c|}{ 9.736E+04} &  \multicolumn{1}{|c|}{ 8.039E+04} \\ 
&$\lvert \dl-d \rvert$ & 
\multicolumn{1}{|c|}{ 2.550E+03} &  \multicolumn{1}{|c|}{ 1.304E+03} &\multicolumn{1}{|c|}{ 8.003E+04} &  \multicolumn{1}{|c|}{ 6.153E+04} \\ 
&$\lvert \de-d \rvert$ & 
\multicolumn{1}{|c|}{ \textbf{ 3.218E+02}} &  \multicolumn{1}{|c|}{ \textbf{ 2.932E+02}} &\multicolumn{1}{|c|}{ \textbf{ 1.632E+04}} &  \multicolumn{1}{|c|}{ \textbf{ 2.559E+04}} \\ 
\hline 

% \multirow{3}{*}{k= 55} 
% &$\lvert \dc-d \rvert$ & 
% \multicolumn{1}{|c|}{ 3.014E+04} &  \multicolumn{1}{|c|}{ 2.947E+04} &\multicolumn{1}{|c|}{ 5.965E+04} &  \multicolumn{1}{|c|}{ 4.522E+04} \\ 
% &$\lvert \dl-d \rvert$ & 
% \multicolumn{1}{|c|}{ 1.872E+03} &  \multicolumn{1}{|c|}{ 9.119E+02} &\multicolumn{1}{|c|}{ 5.009E+04} &  \multicolumn{1}{|c|}{ 3.646E+04} \\ 
% &$\lvert \de-d \rvert$ & 
% \multicolumn{1}{|c|}{ \textbf{ 2.560E+02}} &  \multicolumn{1}{|c|}{ \textbf{ 2.395E+02}} &\multicolumn{1}{|c|}{ \textbf{ 1.075E+04}} &  \multicolumn{1}{|c|}{ \textbf{ 1.486E+04}} \\ 
% \hline 

\multirow{3}{*}{k= 60} 
&$\lvert \dc-d \rvert$ & 
\multicolumn{1}{|c|}{ 1.691E+04} &  \multicolumn{1}{|c|}{ 1.485E+04} &\multicolumn{1}{|c|}{ 3.332E+04} &  \multicolumn{1}{|c|}{ 2.300E+04} \\ 
&$\lvert \dl-d \rvert$ & 
\multicolumn{1}{|c|}{ 1.298E+03} &  \multicolumn{1}{|c|}{ 6.066E+02} &\multicolumn{1}{|c|}{ 2.875E+04} &  \multicolumn{1}{|c|}{ 1.922E+04} \\ 
&$\lvert \de-d \rvert$ & 
\multicolumn{1}{|c|}{ \textbf{ 1.831E+02}} &  \multicolumn{1}{|c|}{ \textbf{ 1.601E+02}} &\multicolumn{1}{|c|}{ \textbf{ 6.158E+03}} &  \multicolumn{1}{|c|}{ \textbf{ 8.067E+03}} \\ 
\hline 

  \end{tabular}
  \end{adjustbox}
\end{table}

%../dataset/wdbc_nolabel.csv
\begin{table}[h] 
\caption{Distance(WDBC)}
\label{tab:DistanceWBCD} 
 
  \begin{adjustbox}{width=\columnwidth} 
  \begin{tabular}{|c|c|c|c|c|c|} 
  \hline 
  
  \multicolumn{2}{|c|}{~} & 
  \multicolumn{2}{|c|}{$x$ and columns of $A$} & 
  \multicolumn{2}{|c|}{column distances} \\ 
  \cline{3-6} 

  \multicolumn{2}{|c|}{~} &mean & std & mean & std \\
  \hline

% \multirow{3}{*}{k= 1} 
% &$\lvert \dc-d \rvert$ & 
% \multicolumn{1}{|c|}{ 1.280E+04} &  \multicolumn{1}{|c|}{ 4.590E+04} &\multicolumn{1}{|c|}{ 2.309E+04} &  \multicolumn{1}{|c|}{ 7.301E+04} \\ 
% &$\lvert \dl-d \rvert$ & 
% \multicolumn{1}{|c|}{ 9.256E+02} &  \multicolumn{1}{|c|}{ 8.296E+02} &\multicolumn{1}{|c|}{ 1.231E+04} &  \multicolumn{1}{|c|}{ 3.108E+04} \\ 
% &$\lvert \de-d \rvert$ & 
% \multicolumn{1}{|c|}{ \textbf{ 1.353E+02}} &  \multicolumn{1}{|c|}{ \textbf{ 1.733E+02}} &\multicolumn{1}{|c|}{ \textbf{ 1.165E+04}} &  \multicolumn{1}{|c|}{ \textbf{ 1.845E+04}} \\ 
% \hline 

\multirow{3}{*}{k= 2} 
&$\lvert \dc-d \rvert$ & 
\multicolumn{1}{|c|}{ 1.980E+03} &  \multicolumn{1}{|c|}{ 1.170E+04} &\multicolumn{1}{|c|}{ 3.522E+03} &  \multicolumn{1}{|c|}{ 1.757E+04} \\ 
&$\lvert \dl-d \rvert$ & 
\multicolumn{1}{|c|}{ 3.478E+02} &  \multicolumn{1}{|c|}{ 3.174E+02} &\multicolumn{1}{|c|}{ 1.878E+03} &  \multicolumn{1}{|c|}{ 4.310E+03} \\ 
&$\lvert \de-d \rvert$ & 
\multicolumn{1}{|c|}{ \textbf{ 5.265E+01}} &  \multicolumn{1}{|c|}{ \textbf{ 6.826E+01}} &\multicolumn{1}{|c|}{ \textbf{ 1.594E+03}} &  \multicolumn{1}{|c|}{ \textbf{ 2.463E+03}} \\ 
\hline 

% \multirow{3}{*}{k= 3} 
% &$\lvert \dc-d \rvert$ & 
% \multicolumn{1}{|c|}{ 6.173E+02} &  \multicolumn{1}{|c|}{ 2.492E+03} &\multicolumn{1}{|c|}{ 1.148E+03} &  \multicolumn{1}{|c|}{ 3.580E+03} \\ 
% &$\lvert \dl-d \rvert$ & 
% \multicolumn{1}{|c|}{ 1.800E+02} &  \multicolumn{1}{|c|}{ 1.805E+02} &\multicolumn{1}{|c|}{ 5.865E+02} &  \multicolumn{1}{|c|}{ 1.410E+03} \\ 
% &$\lvert \de-d \rvert$ & 
% \multicolumn{1}{|c|}{ \textbf{ 2.741E+01}} &  \multicolumn{1}{|c|}{ \textbf{ 3.837E+01}} &\multicolumn{1}{|c|}{ \textbf{ 4.916E+02}} &  \multicolumn{1}{|c|}{ \textbf{ 9.688E+02}} \\ 
% \hline 

\multirow{3}{*}{k= 4} 
&$\lvert \dc-d \rvert$ & 
\multicolumn{1}{|c|}{ 7.563E+01} &  \multicolumn{1}{|c|}{ 7.556E+01} &\multicolumn{1}{|c|}{ 9.894E+01} &  \multicolumn{1}{|c|}{ 1.341E+02} \\ 
&$\lvert \dl-d \rvert$ & 
\multicolumn{1}{|c|}{ 5.945E+01} &  \multicolumn{1}{|c|}{ 4.286E+01} &\multicolumn{1}{|c|}{ 6.877E+01} &  \multicolumn{1}{|c|}{ 1.086E+02} \\ 
&$\lvert \de-d \rvert$ & 
\multicolumn{1}{|c|}{ \textbf{ 9.616E+00}} &  \multicolumn{1}{|c|}{ \textbf{ 1.074E+01}} &\multicolumn{1}{|c|}{ \textbf{ 5.017E+01}} &  \multicolumn{1}{|c|}{ \textbf{ 6.652E+01}} \\ 
\hline 

% \multirow{3}{*}{k= 5} 
% &$\lvert \dc-d \rvert$ & 
% \multicolumn{1}{|c|}{ 3.323E+01} &  \multicolumn{1}{|c|}{ 1.813E+01} &\multicolumn{1}{|c|}{ 1.651E+01} &  \multicolumn{1}{|c|}{ 2.541E+01} \\ 
% &$\lvert \dl-d \rvert$ & 
% \multicolumn{1}{|c|}{ 2.335E+01} &  \multicolumn{1}{|c|}{ 1.776E+01} &\multicolumn{1}{|c|}{ 1.112E+01} &  \multicolumn{1}{|c|}{ 1.704E+01} \\ 
% &$\lvert \de-d \rvert$ & 
% \multicolumn{1}{|c|}{ \textbf{ 3.681E+00}} &  \multicolumn{1}{|c|}{ \textbf{ 4.364E+00}} &\multicolumn{1}{|c|}{ \textbf{ 6.733E+00}} &  \multicolumn{1}{|c|}{ \textbf{ 1.009E+01}} \\ 
% \hline 

\multirow{3}{*}{k= 10} 
&$\lvert \dc-d \rvert$ & 
\multicolumn{1}{|c|}{ 1.990E+01} &  \multicolumn{1}{|c|}{ 6.392E+00} &\multicolumn{1}{|c|}{ 1.044E-01} &  \multicolumn{1}{|c|}{ 1.740E-01} \\ 
&$\lvert \dl-d \rvert$ & 
\multicolumn{1}{|c|}{ 1.627E+00} &  \multicolumn{1}{|c|}{ 1.306E+00} &\multicolumn{1}{|c|}{ 6.845E-02} &  \multicolumn{1}{|c|}{ 1.022E-01} \\ 
&$\lvert \de-d \rvert$ & 
\multicolumn{1}{|c|}{ \textbf{ 2.876E-01}} &  \multicolumn{1}{|c|}{ \textbf{ 3.430E-01}} &\multicolumn{1}{|c|}{ \textbf{ 3.787E-02}} &  \multicolumn{1}{|c|}{ \textbf{ 6.065E-02}} \\ 
\hline 

% \multirow{3}{*}{k= 15} 
% &$\lvert \dc-d \rvert$ & 
% \multicolumn{1}{|c|}{ 1.510E+01} &  \multicolumn{1}{|c|}{ 5.736E+00} &\multicolumn{1}{|c|}{ 4.104E-03} &  \multicolumn{1}{|c|}{ 4.371E-03} \\ 
% &$\lvert \dl-d \rvert$ & 
% \multicolumn{1}{|c|}{ 3.101E-01} &  \multicolumn{1}{|c|}{ 1.884E-01} &\multicolumn{1}{|c|}{ 3.301E-03} &  \multicolumn{1}{|c|}{ 3.031E-03} \\ 
% &$\lvert \de-d \rvert$ & 
% \multicolumn{1}{|c|}{ \textbf{ 6.376E-02}} &  \multicolumn{1}{|c|}{ \textbf{ 6.237E-02}} &\multicolumn{1}{|c|}{ \textbf{ 1.210E-03}} &  \multicolumn{1}{|c|}{ \textbf{ 1.357E-03}} \\ 
% \hline 

\multirow{3}{*}{k= 20} 
&$\lvert \dc-d \rvert$ & 
\multicolumn{1}{|c|}{ 9.954E+00} &  \multicolumn{1}{|c|}{ 4.638E+00} &\multicolumn{1}{|c|}{ 5.085E-04} &  \multicolumn{1}{|c|}{ 5.629E-04} \\ 
&$\lvert \dl-d \rvert$ & 
\multicolumn{1}{|c|}{ 8.621E-02} &  \multicolumn{1}{|c|}{ 6.075E-02} &\multicolumn{1}{|c|}{ 3.935E-04} &  \multicolumn{1}{|c|}{ 3.974E-04} \\ 
&$\lvert \de-d \rvert$ & 
\multicolumn{1}{|c|}{ \textbf{ 2.213E-02}} &  \multicolumn{1}{|c|}{ \textbf{ 2.244E-02}} &\multicolumn{1}{|c|}{ \textbf{ 1.517E-04}} &  \multicolumn{1}{|c|}{ \textbf{ 1.786E-04}} \\ 
\hline 

  \end{tabular}
  \end{adjustbox}
\end{table}

Tables \ref{tab:Distanceionosphere},
\ref{tab:DistanceYearPredictionMSD},
and~\ref{tab:DistanceWBCD}
show the error of computing these distances with the various formulas.
The left part shows the error mean and standard deviation
of the formulas described in Section~\ref{secdax}.
The right part shows the error mean and standard deviation
of the formulas described in Section~\ref{secdaa}.
In all cases
the mean and the standard deviation of the results computed by the
Maximum Entropy Method were the best.

% plots
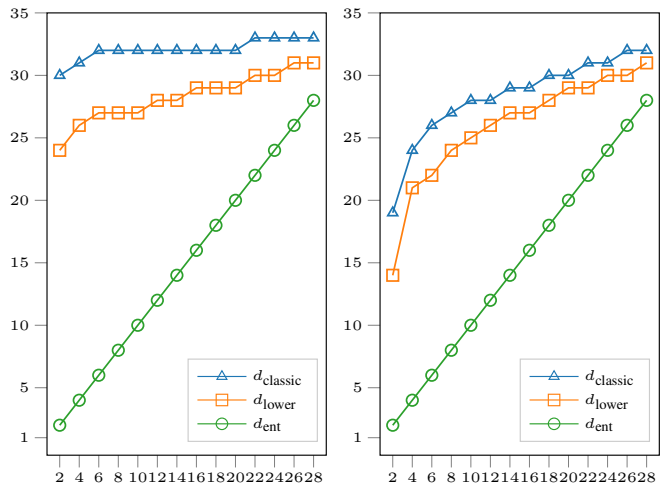
\begin{figure}
  \centering
  \begin{adjustbox}{width=\columnwidth}
    \subfloat{% This file was created by matplotlib2tikz v0.6.18.
\begin{tikzpicture}
\fontsize{5}{5}
\pgfplotsset{width=5cm,height=7cm}

\definecolor{color0}{rgb}{0.12156862745098,0.466666666666667,0.705882352941177}
\definecolor{color1}{rgb}{1,0.498039215686275,0.0549019607843137}
\definecolor{color2}{rgb}{0.172549019607843,0.627450980392157,0.172549019607843}

\begin{axis}[
legend cell align={left},
legend entries={{$\dc$},{$\dl$},{$\de$}},
legend style={at={(0.97,0.03)}, anchor=south east, draw=white!80.0!black},
tick align=outside,
tick pos=left,
x grid style={white!69.01960784313725!black},
xmin=0.7, xmax=29.3,
y grid style={white!69.01960784313725!black},
ymin=-0.4, ymax=35,
xtick={2,4,6,8,10,12,14,16,18,20,22,24,26,28},
ytick={1,5,10,15,20,25,30,35}]
\addlegendimage{mark=triangle, color0}
\addlegendimage{mark=square, color1}
\addlegendimage{mark=o, color2}
\addplot [semithick, color0, mark=triangle, mark size=2, mark options={solid}]
table [row sep=\\]{%
2	30 \\
4	31 \\
6	32 \\
8	32 \\
10	32 \\
12	32 \\
14	32 \\
16	32 \\
18	32 \\
20	32 \\
22	33 \\
24	33 \\
26	33 \\
28	33 \\
};
\addplot [semithick, color1, mark=square, mark size=2, mark options={solid}]
table [row sep=\\]{%
2	24 \\
4	26 \\
6	27 \\
8	27 \\
10	27 \\
12	28 \\
14	28 \\
16	29 \\
18	29 \\
20	29 \\
22	30 \\
24	30 \\
26	31 \\
28	31 \\
};
\addplot [semithick, color2, mark=o, mark size=2, mark options={solid}]
table [row sep=\\]{%
2	2 \\
4	4 \\
6	6 \\
8	8 \\
10	10 \\
12	12 \\
14	14 \\
16	16 \\
18	18 \\
20	20 \\
22	22 \\
24	24 \\
26	26 \\
28	28 \\
};
\end{axis}

\end{tikzpicture}}
    \subfloat{% This file was created by matplotlib2tikz v0.6.18.
\begin{tikzpicture}
\fontsize{5}{5}
\pgfplotsset{width=5cm,height=7cm}

\definecolor{color0}{rgb}{0.12156862745098,0.466666666666667,0.705882352941177}
\definecolor{color1}{rgb}{1,0.498039215686275,0.0549019607843137}
\definecolor{color2}{rgb}{0.172549019607843,0.627450980392157,0.172549019607843}

\begin{axis}[
legend cell align={left},
legend entries={{$\dc$},{$\dl$},{$\de$}},
legend style={at={(0.5,0.03)}, anchor=south west, draw=white!80.0!black},
tick align=outside,
tick pos=left,
x grid style={white!69.01960784313725!black},
xmin=0.7, xmax=29.3,
y grid style={white!69.01960784313725!black},
ymin=-0.4, ymax=35,
xtick={2,4,6,8,10,12,14,16,18,20,22,24,26,28},
ytick={1,5,10,15,20,25,30,35}]
\addlegendimage{mark=triangle, color0}
\addlegendimage{mark=square, color1}
\addlegendimage{mark=o, color2}
\addplot [semithick, color0, mark=triangle, mark size=2, mark options={solid}]
table [row sep=\\]{%
2	19 \\
4	24 \\
6	26 \\
8	27 \\
10	28 \\
12	28 \\
14	29 \\
16	29 \\
18	30 \\
20	30 \\
22	31 \\
24	31 \\
26	32 \\
28	32 \\
};
\addplot [semithick, color1, mark=square, mark size=2, mark options={solid}]
table [row sep=\\]{%
2	14 \\
4	21 \\
6	22 \\
8	24 \\
10	25 \\
12	26 \\
14	27 \\
16	27 \\
18	28 \\
20	29 \\
22	29 \\
24	30 \\
26	30 \\
28	31 \\
};
\addplot [semithick, color2, mark=o, mark size=2, mark options={solid}]
table [row sep=\\]{%
2	2 \\
4	4 \\
6	6 \\
8	8 \\
10	10 \\
12	12 \\
14	14 \\
16	16 \\
18	18 \\
20	20 \\
22	22 \\
24	24 \\
26	26 \\
28	28 \\
};
\end{axis}

\end{tikzpicture}}
  \end{adjustbox}
  \caption{
    Comparision of $k$  with a fixed error value.
    Dataset: Ionosphere.
    Left panel: Distance between $x$ and the columns of $A$;
    Right panel: Column distances.
  }
  \label{fig:distanceionosphere}
\end{figure}

\begin{figure}
  \centering
  \begin{adjustbox}{width=\columnwidth}
    \subfloat{% This file was created by matplotlib2tikz v0.6.18.
\begin{tikzpicture}
\fontsize{5}{5}
\pgfplotsset{width=5cm,height=7cm}

\definecolor{color0}{rgb}{0.12156862745098,0.466666666666667,0.705882352941177}
\definecolor{color1}{rgb}{1,0.498039215686275,0.0549019607843137}
\definecolor{color2}{rgb}{0.172549019607843,0.627450980392157,0.172549019607843}

\begin{axis}[
legend cell align={left},
legend entries={{$\dc$},{$\dl$},{$\de$}},
legend style={at={(0.97,0.03)}, anchor=south east, draw=white!80.0!black},
tick align=outside,
tick pos=left,
x grid style={white!69.01960784313725!black},
xmin=-1.95, xmax=62.95,
y grid style={white!69.01960784313725!black},
ymin=-0.4, ymax=87.1,
xtick={1,5,10,15,20,25,30,35,40,45,50,55,60},
ytick={1,5,10,15,20,25,30,35,40,45,50,55,60,65,70,75,80,85}]
\addlegendimage{mark=triangle, color0}
\addlegendimage{mark=square, color1}
\addlegendimage{mark=o, color2}
\addplot [semithick, color0, mark=triangle, mark size=2, mark options={solid}]
table [row sep=\\]{%
1	73 \\
5	76 \\
10	77 \\
15	78 \\
20	79 \\
25	79 \\
30	80 \\
35	80 \\
40	81 \\
45	81 \\
50	82 \\
55	83 \\
60	83 \\
};
\addplot [semithick, color1, mark=square, mark size=2, mark options={solid}]
table [row sep=\\]{%
1	51 \\
5	57 \\
10	61 \\
15	63 \\
20	65 \\
25	68 \\
30	70 \\
35	71 \\
40	73 \\
45	75 \\
50	76 \\
55	77 \\
60	79 \\
};
\addplot [semithick, color2, mark=o, mark size=2, mark options={solid}]
table [row sep=\\]{%
1	1 \\
5	5 \\
10	10 \\
15	15 \\
20	20 \\
25	25 \\
30	30 \\
35	35 \\
40	40 \\
45	45 \\
50	50 \\
55	55 \\
60	60 \\
};
\end{axis}

\end{tikzpicture}}
    \subfloat{% This file was created by matplotlib2tikz v0.6.18.
\begin{tikzpicture}
\fontsize{5}{5}
\pgfplotsset{width=5cm,height=7cm}

\definecolor{color0}{rgb}{0.12156862745098,0.466666666666667,0.705882352941177}
\definecolor{color1}{rgb}{1,0.498039215686275,0.0549019607843137}
\definecolor{color2}{rgb}{0.172549019607843,0.627450980392157,0.172549019607843}

\begin{axis}[
legend cell align={left},
legend entries={{$\dc$},{$\dl$},{$\de$}},
legend style={at={(0.03,0.97)}, anchor=north west, draw=white!80.0!black},
tick align=outside,
tick pos=left,
x grid style={white!69.01960784313725!black},
xmin=-1.95, xmax=62.95,
y grid style={white!69.01960784313725!black},
ymin=-0.4, ymax=85,
xtick={1,5,10,15,20,25,30,35,40,45,50,55,60},
ytick={1,5,10,15,20,25,30,35,40,45,50,55,60,65,70,75,80,85}]
\addlegendimage{mark=triangle, color0}
\addlegendimage{mark=square, color1}
\addlegendimage{mark=o, color2}
\addplot [semithick, color0, mark=triangle, mark size=2, mark options={solid}]
table [row sep=\\]{%
1	11 \\
5	26 \\
10	36 \\
15	44 \\
20	48 \\
25	51 \\
30	54 \\
35	57 \\
40	60 \\
45	63 \\
50	66 \\
55	70 \\
60	72 \\
};
\addplot [semithick, color1, mark=square, mark size=2, mark options={solid}]
table [row sep=\\]{%
1	13 \\
5	25 \\
10	32 \\
15	38 \\
20	42 \\
25	49 \\
30	53 \\
35	56 \\
40	59 \\
45	62 \\
50	66 \\
55	69 \\
60	71 \\
};
\addplot [semithick, color2, mark=o, mark size=2, mark options={solid}]
table [row sep=\\]{%
1	1 \\
5	5 \\
10	10 \\
15	15 \\
20	20 \\
25	25 \\
30	30 \\
35	35 \\
40	40 \\
45	45 \\
50	50 \\
55	55 \\
60	60 \\
};
\end{axis}

\end{tikzpicture}}
  \end{adjustbox}
  \caption{
    Comparision of $k$  with a fixed error value.
    Dataset: YearPredictionMSD.
    Left panel: Distance between $x$ and the columns of $A$;
    Right panel: Column distances.
  }
  \label{fig:distancemsd}
\end{figure}
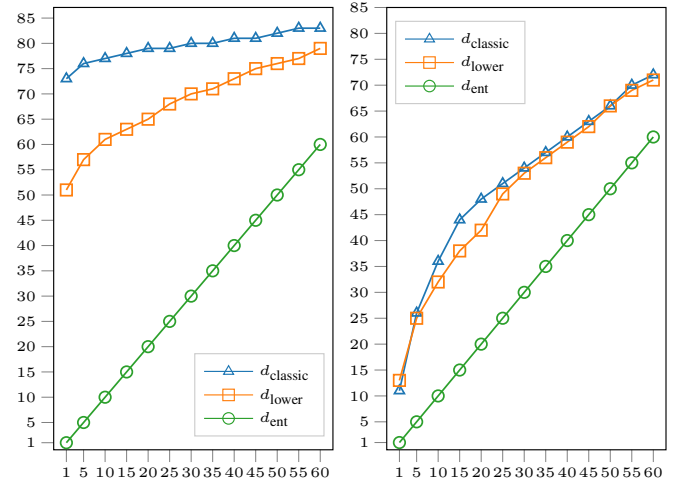

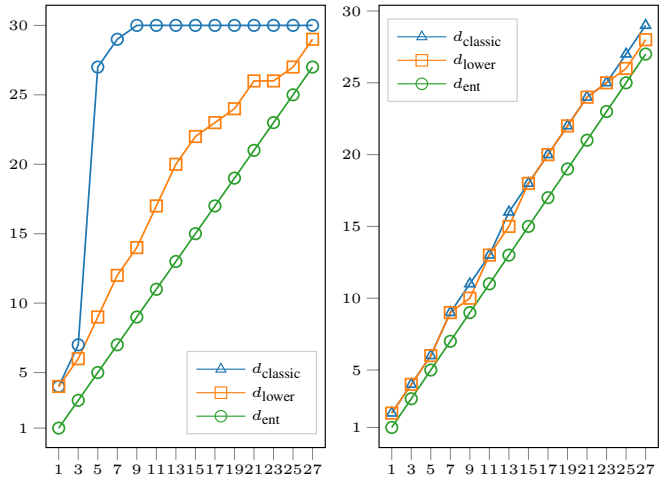
\begin{figure}
  \centering
  \begin{adjustbox}{width=\columnwidth}
    \subfloat{% This file was created by matplotlib2tikz v0.6.18.
\begin{tikzpicture}
\fontsize{5}{5}
\pgfplotsset{width=5cm,height=7cm}

\definecolor{color0}{rgb}{0.12156862745098,0.466666666666667,0.705882352941177}
\definecolor{color1}{rgb}{1,0.498039215686275,0.0549019607843137}
\definecolor{color2}{rgb}{0.172549019607843,0.627450980392157,0.172549019607843}

\begin{axis}[
legend cell align={left},
legend entries={{$\dc$},{$\dl$},{$\de$}},
legend style={at={(0.97,0.03)}, anchor=south east, draw=white!80.0!black},
tick align=outside,
tick pos=left,
x grid style={white!69.01960784313725!black},
xmin=-0.3, xmax=28.3,
y grid style={white!69.01960784313725!black},
ymin=-0.4, ymax=31.45,
xtick={1,3,5,7,9,11,13,15,17,19,21,23,25,27},
ytick={1,5,10,15,20,25,30}]
\addlegendimage{mark=triangle, color0}
\addlegendimage{mark=square, color1}
\addlegendimage{mark=o, color2}
\addplot [semithick, color0, mark=o, mark size=2, mark options={solid}]
table [row sep=\\]{%
1	4 \\
3	7 \\
5	27 \\
7	29 \\
9	30 \\
11	30 \\
13	30 \\
15	30 \\
17	30 \\
19	30 \\
21	30 \\
23	30 \\
25	30 \\
27	30 \\
};
\addplot [semithick, color1, mark=square, mark size=2, mark options={solid}]
table [row sep=\\]{%
1	4 \\
3	6 \\
5	9 \\
7	12 \\
9	14 \\
11	17 \\
13	20 \\
15	22 \\
17	23 \\
19	24 \\
21	26 \\
23	26 \\
25	27 \\
27	29 \\
};
\addplot [semithick, color2, mark=o, mark size=2, mark options={solid}]
table [row sep=\\]{%
1	1 \\
3	3 \\
5	5 \\
7	7 \\
9	9 \\
11	11 \\
13	13 \\
15	15 \\
17	17 \\
19	19 \\
21	21 \\
23	23 \\
25	25 \\
27	27 \\
};
\end{axis}

\end{tikzpicture}}
    \subfloat{% This file was created by matplotlib2tikz v0.6.18.
\begin{tikzpicture}
\fontsize{5}{5}
\pgfplotsset{width=5cm,height=7cm}

\definecolor{color0}{rgb}{0.12156862745098,0.466666666666667,0.705882352941177}
\definecolor{color1}{rgb}{1,0.498039215686275,0.0549019607843137}
\definecolor{color2}{rgb}{0.172549019607843,0.627450980392157,0.172549019607843}

\begin{axis}[
legend cell align={left},
legend entries={{$\dc$},{$\dl$},{$\de$}},
legend style={at={(0.03,0.97)}, anchor=north west, draw=white!80.0!black},
tick align=outside,
tick pos=left,
x grid style={white!69.01960784313725!black},
xmin=-0.3, xmax=28.3,
y grid style={white!69.01960784313725!black},
ymin=-0.4, ymax=30.4,
xtick={1,3,5,7,9,11,13,15,17,19,21,23,25,27},
ytick={1,5,10,15,20,25,30}]
\addlegendimage{mark=triangle, color0}
\addlegendimage{mark=square, color1}
\addlegendimage{mark=o, color2}
\addplot [semithick, color0, mark=triangle, mark size=2, mark options={solid}]
table [row sep=\\]{%
1	2 \\
3	4 \\
5	6 \\
7	9 \\
9	11 \\
11	13 \\
13	16 \\
15	18 \\
17	20 \\
19	22 \\
21	24 \\
23	25 \\
25	27 \\
27	29 \\
};
\addplot [semithick, color1, mark=square, mark size=2, mark options={solid}]
table [row sep=\\]{%
1	2 \\
3	4 \\
5	6 \\
7	9 \\
9	10 \\
11	13 \\
13	15 \\
15	18 \\
17	20 \\
19	22 \\
21	24 \\
23	25 \\
25	26 \\
27	28 \\
};
\addplot [semithick, color2, mark=o, mark size=2, mark options={solid}]
table [row sep=\\]{%
1	1 \\
3	3 \\
5	5 \\
7	7 \\
9	9 \\
11	11 \\
13	13 \\
15	15 \\
17	17 \\
19	19 \\
21	21 \\
23	23 \\
25	25 \\
27	27 \\
};
\end{axis}

\end{tikzpicture}}
  \end{adjustbox}
  \caption{
    Comparision of $k$  with a fixed error value.
    Dataset: wdbc.
    Left panel: Distance between $x$ and the columns of $A$;
    Right panel: Column distances.
  }
  \label{fig:distancewdbc}
\end{figure}

To quantify  the advantage of the new formulas over the classical formulas
we ran the following set of
experiments. For a fixed value of $k$ the formula $\de$ was applied to the data
 and its error was measured.
 We then applied $\dc$ and $\dl$ to the same data, and increased the value of
 $k$ until they produced the same error.
 The results for different datasets are shown in figures 
\ref{fig:distanceionosphere},
\ref{fig:distancemsd},
and \ref{fig:distancewdbc}.
For example,  Fig.\ref{fig:distanceionosphere} was computed for the Ionosphare
dataset. 
To obtain the same error of $\de$ with $k=2$ the formula $\dl$ needs $k=24$,
and the formula $\dc$ needs 30.
We observe that the advantage of $\de$ over $\dl$ and $\dc$ is quite
significant, for the ``Ionosphare'' and the ``YearPredictioMSD'' datasets.
They are not that impressive for the ``wdbc'' dataset.

\subsection{Experiments with Rayleigh Quotients}

% tables

 %../dataset/ionosphere_nolabel.csv
\begin{table}[h]
 \caption{Rayleigh Quotient(Ionosphere)
 }
 \label{tab:RQ1}
 \begin{adjustbox}{width=\columnwidth}
  \begin{tabular}{|c|c|c|c|c|c|}
  \hline

  \multicolumn{2}{|c|}{~} &
  \multicolumn{2}{|c|}{Column} & 
  \multicolumn{2}{|c|}{Row}\\
  \cline{3-6}

  \multicolumn{2}{|c|}{~} &mean & std & mean & std\\
  \hline

\multirow{2}{*}{k= 2}
&$\lvert \rc-r \rvert$ &
\multicolumn{1}{|c|}{ 6.151E+01} &  \multicolumn{1}{|c|}{ 1.800E+01} &
\multicolumn{1}{|c|}{ 5.733E+00} &  \multicolumn{1}{|c|}{ 2.098E+00} \\
&$\lvert \re-r \rvert$ &
\multicolumn{1}{|c|}{ \textbf{ 1.401E+01}} &  \multicolumn{1}{|c|}{ \textbf{ 9.829E+00}} &
\multicolumn{1}{|c|}{ \textbf{ 1.475E+00}} &  \multicolumn{1}{|c|}{ \textbf{ 1.271E+00}} \\
\hline

% \multirow{2}{*}{k= 4}
% &$\lvert \rc-r \rvert$ &
% \multicolumn{1}{|c|}{ 4.231E+01} &  \multicolumn{1}{|c|}{ 7.649E+00} &
% \multicolumn{1}{|c|}{ 3.998E+00} &  \multicolumn{1}{|c|}{ 9.894E-01} \\
% &$\lvert \re-r \rvert$ &
% \multicolumn{1}{|c|}{ \textbf{ 4.821E+00}} &  \multicolumn{1}{|c|}{ \textbf{ 4.111E+00}} &
% \multicolumn{1}{|c|}{ \textbf{ 7.682E-01}} &  \multicolumn{1}{|c|}{ \textbf{ 5.530E-01}} \\
% \hline

\multirow{2}{*}{k= 6}
&$\lvert \rc-r \rvert$ &
\multicolumn{1}{|c|}{ 3.507E+01} &  \multicolumn{1}{|c|}{ 6.625E+00} &
\multicolumn{1}{|c|}{ 3.074E+00} &  \multicolumn{1}{|c|}{ 8.803E-01} \\
&$\lvert \re-r \rvert$ &
\multicolumn{1}{|c|}{ \textbf{ 4.152E+00}} &  \multicolumn{1}{|c|}{ \textbf{ 2.999E+00}} &
\multicolumn{1}{|c|}{ \textbf{ 6.486E-01}} &  \multicolumn{1}{|c|}{ \textbf{ 4.849E-01}} \\
\hline

% \multirow{2}{*}{k= 8}
% &$\lvert \rc-r \rvert$ &
% \multicolumn{1}{|c|}{ 2.977E+01} &  \multicolumn{1}{|c|}{ 5.973E+00} &
% \multicolumn{1}{|c|}{ 2.722E+00} &  \multicolumn{1}{|c|}{ 8.099E-01} \\
% &$\lvert \re-r \rvert$ &
% \multicolumn{1}{|c|}{ \textbf{ 3.757E+00}} &  \multicolumn{1}{|c|}{ \textbf{ 2.780E+00}} &
% \multicolumn{1}{|c|}{ \textbf{ 5.700E-01}} &  \multicolumn{1}{|c|}{ \textbf{ 4.532E-01}} \\
% \hline

\multirow{2}{*}{k= 10}
&$\lvert \rc-r \rvert$ &
\multicolumn{1}{|c|}{ 2.383E+01} &  \multicolumn{1}{|c|}{ 3.688E+00} &
\multicolumn{1}{|c|}{ 2.370E+00} &  \multicolumn{1}{|c|}{ 9.197E-01} \\
&$\lvert \re-r \rvert$ &
\multicolumn{1}{|c|}{ \textbf{ 2.125E+00}} &  \multicolumn{1}{|c|}{ \textbf{ 1.431E+00}} &
\multicolumn{1}{|c|}{ \textbf{ 6.627E-01}} &  \multicolumn{1}{|c|}{ \textbf{ 5.844E-01}} \\
\hline

% \multirow{2}{*}{k= 12}
% &$\lvert \rc-r \rvert$ &
% \multicolumn{1}{|c|}{ 1.896E+01} &  \multicolumn{1}{|c|}{ 4.413E+00} &
% \multicolumn{1}{|c|}{ 1.758E+00} &  \multicolumn{1}{|c|}{ 6.157E-01} \\
% &$\lvert \re-r \rvert$ &
% \multicolumn{1}{|c|}{ \textbf{ 2.367E+00}} &  \multicolumn{1}{|c|}{ \textbf{ 1.571E+00}} &
% \multicolumn{1}{|c|}{ \textbf{ 4.523E-01}} &  \multicolumn{1}{|c|}{ \textbf{ 3.163E-01}} \\
% \hline

\multirow{2}{*}{k= 14}
&$\lvert \rc-r \rvert$ &
\multicolumn{1}{|c|}{ 1.423E+01} &  \multicolumn{1}{|c|}{ 4.056E+00} &
\multicolumn{1}{|c|}{ 1.401E+00} &  \multicolumn{1}{|c|}{ 3.911E-01} \\
&$\lvert \re-r \rvert$ &
\multicolumn{1}{|c|}{ \textbf{ 1.519E+00}} &  \multicolumn{1}{|c|}{ \textbf{ 1.103E+00}} &
\multicolumn{1}{|c|}{ \textbf{ 3.334E-01}} &  \multicolumn{1}{|c|}{ \textbf{ 2.417E-01}} \\
\hline

% \multirow{2}{*}{k= 16}
% &$\lvert \rc-r \rvert$ &
% \multicolumn{1}{|c|}{ 1.288E+01} &  \multicolumn{1}{|c|}{ 3.432E+00} &
% \multicolumn{1}{|c|}{ 1.184E+00} &  \multicolumn{1}{|c|}{ 3.902E-01} \\
% &$\lvert \re-r \rvert$ &
% \multicolumn{1}{|c|}{ \textbf{ 1.372E+00}} &  \multicolumn{1}{|c|}{ \textbf{ 1.024E+00}} &
% \multicolumn{1}{|c|}{ \textbf{ 2.860E-01}} &  \multicolumn{1}{|c|}{ \textbf{ 2.220E-01}} \\
% \hline

\multirow{2}{*}{k= 18}
&$\lvert \rc-r \rvert$ &
\multicolumn{1}{|c|}{ 9.712E+00} &  \multicolumn{1}{|c|}{ 2.699E+00} &
\multicolumn{1}{|c|}{ 9.291E-01} &  \multicolumn{1}{|c|}{ 3.089E-01} \\
&$\lvert \re-r \rvert$ &
\multicolumn{1}{|c|}{ \textbf{ 1.139E+00}} &  \multicolumn{1}{|c|}{ \textbf{ 8.474E-01}} &
\multicolumn{1}{|c|}{ \textbf{ 2.149E-01}} &  \multicolumn{1}{|c|}{ \textbf{ 1.376E-01}} \\
\hline

% \multirow{2}{*}{k= 20}
% &$\lvert \rc-r \rvert$ &
% \multicolumn{1}{|c|}{ 8.454E+00} &  \multicolumn{1}{|c|}{ 2.516E+00} &
% \multicolumn{1}{|c|}{ 7.817E-01} &  \multicolumn{1}{|c|}{ 3.233E-01} \\
% &$\lvert \re-r \rvert$ &
% \multicolumn{1}{|c|}{ \textbf{ 8.226E-01}} &  \multicolumn{1}{|c|}{ \textbf{ 6.997E-01}} &
% \multicolumn{1}{|c|}{ \textbf{ 2.639E-01}} &  \multicolumn{1}{|c|}{ \textbf{ 1.904E-01}} \\
% \hline

\multirow{2}{*}{k= 22}
&$\lvert \rc-r \rvert$ &
\multicolumn{1}{|c|}{ 5.922E+00} &  \multicolumn{1}{|c|}{ 2.281E+00} &
\multicolumn{1}{|c|}{ 6.001E-01} &  \multicolumn{1}{|c|}{ 2.420E-01} \\
&$\lvert \re-r \rvert$ &
\multicolumn{1}{|c|}{ \textbf{ 6.551E-01}} &  \multicolumn{1}{|c|}{ \textbf{ 6.374E-01}} &
\multicolumn{1}{|c|}{ \textbf{ 1.618E-01}} &  \multicolumn{1}{|c|}{ \textbf{ 1.417E-01}} \\
\hline

% \multirow{2}{*}{k= 24}
% &$\lvert \rc-r \rvert$ &
% \multicolumn{1}{|c|}{ 4.678E+00} &  \multicolumn{1}{|c|}{ 1.826E+00} &
% \multicolumn{1}{|c|}{ 4.253E-01} &  \multicolumn{1}{|c|}{ 1.434E-01} \\
% &$\lvert \re-r \rvert$ &
% \multicolumn{1}{|c|}{ \textbf{ 3.296E-01}} &  \multicolumn{1}{|c|}{ \textbf{ 3.468E-01}} &
% \multicolumn{1}{|c|}{ \textbf{ 1.110E-01}} &  \multicolumn{1}{|c|}{ \textbf{ 9.019E-02}} \\
% \hline

\multirow{2}{*}{k= 26}
&$\lvert \rc-r \rvert$ &
\multicolumn{1}{|c|}{ 3.007E+00} &  \multicolumn{1}{|c|}{ 1.276E+00} &
\multicolumn{1}{|c|}{ 3.354E-01} &  \multicolumn{1}{|c|}{ 2.059E-01} \\
&$\lvert \re-r \rvert$ &
\multicolumn{1}{|c|}{ \textbf{ 2.995E-01}} &  \multicolumn{1}{|c|}{ \textbf{ 2.960E-01}} &
\multicolumn{1}{|c|}{ \textbf{ 1.353E-01}} &  \multicolumn{1}{|c|}{ \textbf{ 1.416E-01}} \\
\hline

% \multirow{2}{*}{k= 28}
% &$\lvert \rc-r \rvert$ &
% \multicolumn{1}{|c|}{ 1.833E+00} &  \multicolumn{1}{|c|}{ 1.169E+00} &
% \multicolumn{1}{|c|}{ 1.593E-01} &  \multicolumn{1}{|c|}{ 1.037E-01} \\
% &$\lvert \re-r \rvert$ &
% \multicolumn{1}{|c|}{ \textbf{ 1.780E-01}} &  \multicolumn{1}{|c|}{ \textbf{ 1.527E-01}} &
% \multicolumn{1}{|c|}{ \textbf{ 8.354E-02}} &  \multicolumn{1}{|c|}{ \textbf{ 5.995E-02}} \\
% \hline

 \end{tabular}
 \end{adjustbox}
\end{table}

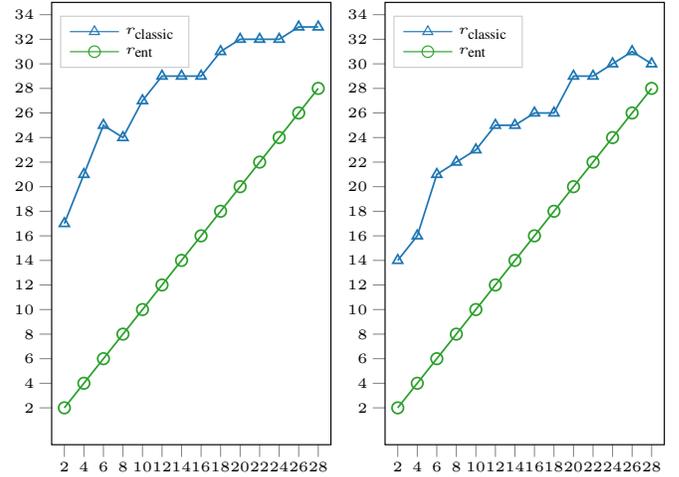
\begin{figure}
  \centering
  \begin{adjustbox}{width=\columnwidth}
    \subfloat{% This file was created by matplotlib2tikz v0.6.18.
\begin{tikzpicture}
\fontsize{5}{5}
\pgfplotsset{width=5cm,height=7cm}
\definecolor{color0}{rgb}{0.12156862745098,0.466666666666667,0.705882352941177}
\definecolor{color1}{rgb}{0.172549019607843,0.627450980392157,0.172549019607843}

\begin{axis}[
legend cell align={left},
legend entries={{$\rc$},{$\re$}},
legend style={at={(0.03,0.97)}, anchor=north west, draw=white!80.0!black},
tick align=outside,
tick pos=left,
x grid style={white!69.01960784313725!black},
xmin=0.7, xmax=29.3,
y grid style={white!69.01960784313725!black},
ymin=-1.0, ymax=35,
xtick={2,4,6,8,10,12,14,16,18,20,22,24,26,28,30,32,34},
ytick={2,4,6,8,10,12,14,16,18,20,22,24,26,28,30,32,34}]
\addlegendimage{mark=triangle, color0}
\addlegendimage{mark=o, color1}
\addplot [semithick, color0, mark=triangle, mark size=2, mark options={solid}]
table [row sep=\\]{%
2	17 \\
4	21 \\
6	25 \\
8	24 \\
10	27 \\
12	29 \\
14	29 \\
16	29 \\
18	31 \\
20	32 \\
22	32 \\
24	32 \\
26	33 \\
28	33 \\
};
\addplot [semithick, color1, mark=o, mark size=2, mark options={solid}]
table [row sep=\\]{%
2	2 \\
4	4 \\
6	6 \\
8	8 \\
10	10 \\
12	12 \\
14	14 \\
16	16 \\
18	18 \\
20	20 \\
22	22 \\
24	24 \\
26	26 \\
28	28 \\
};
\end{axis}

\end{tikzpicture}}
    \subfloat{% This file was created by matplotlib2tikz v0.6.18.
\begin{tikzpicture}
\fontsize{5}{5}
\pgfplotsset{width=5cm,height=7cm}
\definecolor{color0}{rgb}{0.12156862745098,0.466666666666667,0.705882352941177}
\definecolor{color1}{rgb}{0.172549019607843,0.627450980392157,0.172549019607843}

\begin{axis}[
legend cell align={left},
legend entries={{$\rc$},{$\re$}},
legend style={at={(0.03,0.97)}, anchor=north west, draw=white!80.0!black},
tick align=outside,
tick pos=left,
x grid style={white!69.01960784313725!black},
xmin=0.7, xmax=29.3,
y grid style={white!69.01960784313725!black},
ymin=-1.0, ymax=35,
xtick={2,4,6,8,10,12,14,16,18,20,22,24,26,28,30,32,34},
ytick={2,4,6,8,10,12,14,16,18,20,22,24,26,28,30,32,34}]
\addlegendimage{mark=triangle, color0}
\addlegendimage{mark=o, color1}
\addplot [semithick, color0, mark=triangle, mark size=2, mark options={solid}]
table [row sep=\\]{%
2	14 \\
4	16 \\
6	21 \\
8	22 \\
10	23 \\
12	25 \\
14	25 \\
16	26 \\
18	26 \\
20	29 \\
22	29 \\
24	30 \\
26	31 \\
28	30 \\
};
\addplot [semithick, color1, mark=o, mark size=2, mark options={solid}]
table [row sep=\\]{%
2	2 \\
4	4 \\
6	6 \\
8	8 \\
10	10 \\
12	12 \\
14	14 \\
16	16 \\
18	18 \\
20	20 \\
22	22 \\
24	24 \\
26	26 \\
28	28 \\
};
\end{axis}

\end{tikzpicture}}
  \end{adjustbox}
  \caption{
    Comparision of $k$ values with fixed error value.
    Dataset: Ionosphere.
    Left panel: Column space Rayleigh Quotient; 
    Right panel: Row space Rayleigh Quotient.
  }
  \label{fig:rq1}
\end{figure}

Table~\ref{tab:RQ1} describes the average difference 
in evaluating the column and row space Rayleigh quotient.
Smaller mean and standard deviation of $\lvert \re-r \rvert$ indicate 
better estimates for the Maximum Entropy Method.
The vectors evaluated in this experiment were randomly drawn
from Gaussian distribution.
The plots in~\ref{fig:rq1} show the advantage of the new formulas using the
same format as in
\ref{fig:distanceionosphere},
\ref{fig:distancemsd},
and \ref{fig:distancewdbc}.

% plots

\bsection{Concluding remarks}

This paper considers a common situation in which
a matrix $A$ is approximated by PCA as: $A =VW$.
A nice aspect of this representation
is that a lot of the operations that involve matrix data can be performed
``in the PCA space'', without reconstructing the matrix or any of its columns.
The paper discusses two of this cases.
The first is computing distances that involve matrix columns,
and the second is the computation of Rayleigh quotients.

Our main result is a novel method of modeling the uncertainty in
the estimates that one obtains from PCA approximations.
The idea is to replace the unknown quantities with random variables.
Using information that is typically available during the creation of
the matrix $W$ in the above estimation and the Maximum Entropy Method
one can determine the likely distribution of these random variables.
Thus, evaluating expressions that
involve the matrix $A$ become estimates of
expected values.

Applying this framework allows us to derive closed form solutions
to distances and Rayleigh quotients that appear to be novel.
Experimental results show that these new formulas produce a significant
improvement in accuracy, when compared to the classical formulas.

\bibliography{articles,books,inproc,rest}
\bibliographystyle{IEEEtran}

\end{document}